\documentclass[
    a4paper,          % Tamanho da folha A4
    12pt,             % Tamanho da fonte 12pt
    chapter=TITLE,    % Todos os capitulos devem ter caixa alta
    section=TITLE,    % Todas as secoes devem ter caixa alta
    oneside,          % Usada para impressao em apenas uma face do papel
    english,          % Hifenizacoes em ingles
    spanish,          % Hifenizacoes em espanhol
    brazil            % Ultimo idioma eh o idioma padrao do documento
]{abntex2}

\usepackage[brazil]{babel} 
\usepackage[utf8]{inputenc}                         % Acentuação direta
\usepackage[T1]{fontenc}                            % Codificação da fonte em 8 bits
\usepackage{graphicx}                               % Inserir figuras
\usepackage{amsfonts, amssymb, amsmath}             % Fonte e símbolos matemáticos
\usepackage{booktabs}                               % Comandos para tabelas
\usepackage{verbatim}                               % Texto é interpretado como escrito no documento
\usepackage{multirow, array}                        % Múltiplas linhas e colunas em tabelas
\usepackage{indentfirst}                            % Endenta o primeiro parágrafo de cada seção.
\usepackage{listings}                               % Utilizar codigo fonte no documento
\usepackage{xcolor}
\usepackage{microtype}                              % Para melhorias de justificação?
\usepackage[portuguese,ruled,lined]{algorithm2e}    % Escrever algoritmos
\usepackage{algorithmic}                            % Criar Algoritmos
\usepackage{amsgen}
\usepackage{lipsum}                                 % Usar a simulação de texto Lorem Ipsum
\usepackage{tocloft}                                % Permite alterar a formatação do Sumário
\usepackage{etoolbox}                               % Usado para alterar a fonte da Section no Sumário
\usepackage[nogroupskip,nonumberlist,acronym]{glossaries}                % Permite fazer o glossario
\usepackage{caption}                                % Altera o comportamento da tag caption
\usepackage[alf, abnt-emphasize=bf, bibjustif, recuo=0cm, abnt-etal-cite=3, abnt-etal-list=0,abnt-etal-text=it]{abntex2cite}  % Citações padrão ABNT
\usepackage{mathptmx}                               % Usa a fonte Times New Roman
\usepackage{appendix}                               % Gerar o apendice no final do documento
\usepackage{paracol}                                % Criar paragrafos sem identacao
\usepackage{lib/unifortex2}		                    % Biblioteca com as normas da Unifor para trabalhos academicos
\usepackage{pdfpages}                               % Incluir pdf no documento
\usepackage{amsmath}                                % Usar equacoes matematicas

\makeglossaries

\makeindex

\trabalhoacademico{tccgraduacao}

\removerbordasdohyperlink{sim}

\cordohyperlink{nao}

\ies{Universidade de Fortaleza}
\iessigla{UNIFOR}
\centro{Centro de Ciências Tecnológicas}

\graduacaoem{Ciência da Computação}
\habilitacao{bacharel} % Pode colocar tambem 'licenciada'

\autor{Daniel Aragão Abreu Filho}
\titulo{Busca de melhor caminho entre múltiplas origens e múltiplos destinos em redes complexas que representam cidades}
\data{2019}
\local{Fortaleza -- Ceará}

\dataaprovacao{09 de Dezembro de 2019}

\orientador{Carlos Caminha}
\orientadories{Universidade de Fortaleza - UNIFOR}
\orientadorcentro{Centro de Ciências Tecnológicas - CCT}
\orientadorfeminino{nao} % Coloque 'sim' se for do sexo feminino

\membrodabancadois{Luciano Heitor Gallegos Marin}
\membrodabancadoisies{Universidade de Fortaleza - Unifor}

\membrodabancatres{Caio Cesar Ponte Silva }
\membrodabancatresies{Universidade de Fortaleza - Unifor}

\begin{document}

	% Elementos pré-textuais
	\imprimircapa
	%\imprimirfolhaderosto{}
	\imprimirfichacatalografica{elementos-pre-textuais/ficha-catalografica}
	%\imprimirerrata{elementos-pre-textuais/errata}

	%%%%%%%%%%%%%%%%%%%%%%%%%%%%%%%%%%%%%%%%%%%%%%%%%%%%%%%%%%%%%%%%%%%%%%%%%%%%%%%%%%%%%%%%%%%%%%%%%%
	%%																                               	%%
	%%		Após a defesa do TCC, a banca examinadora irá preencher a folha de aprovação,			%%
	%%		que deverá ser assinada por todos os membros da banca avaliadora e, posteriormente,		%%
	%%		deverá ser anexada na versão final do TCC pelo ALUNO.									%%
	%%																								%%
	%%		Quando tiver em mãos a folha de aprovação:												%%
	%%			1 - Escanei-a e gere um arquivo PDF													%%
	%%			2 - Renomeie o arquivo PDF para "folha-aprovacao.pdf"								%%
	%%			3 - Mova o arquivo para o diretório "elementos-pre-textuais" do modelo				%%
	%%			5 - Descomente a linha \includepdf{elementos-pre-textuais/folha-aprovacao.pdf} 		%%
	%%					removendo o simbolo de %													%%
	%%			6 - Adicione o simbolo % na linha 	\imprimirfolhadeaprovacao						%%
	%%			7 - Recompile o projeto																%%
	%%																								%%	%%%%%%%%%%%%%%%%%%%%%%%%%%%%%%%%%%%%%%%%%%%%%%%%%%%%%%%%%%%%%%%%%%%%%%%%%%%%%%%%%%%%%%%%%%%%%%%%%%
	
    \includepdf{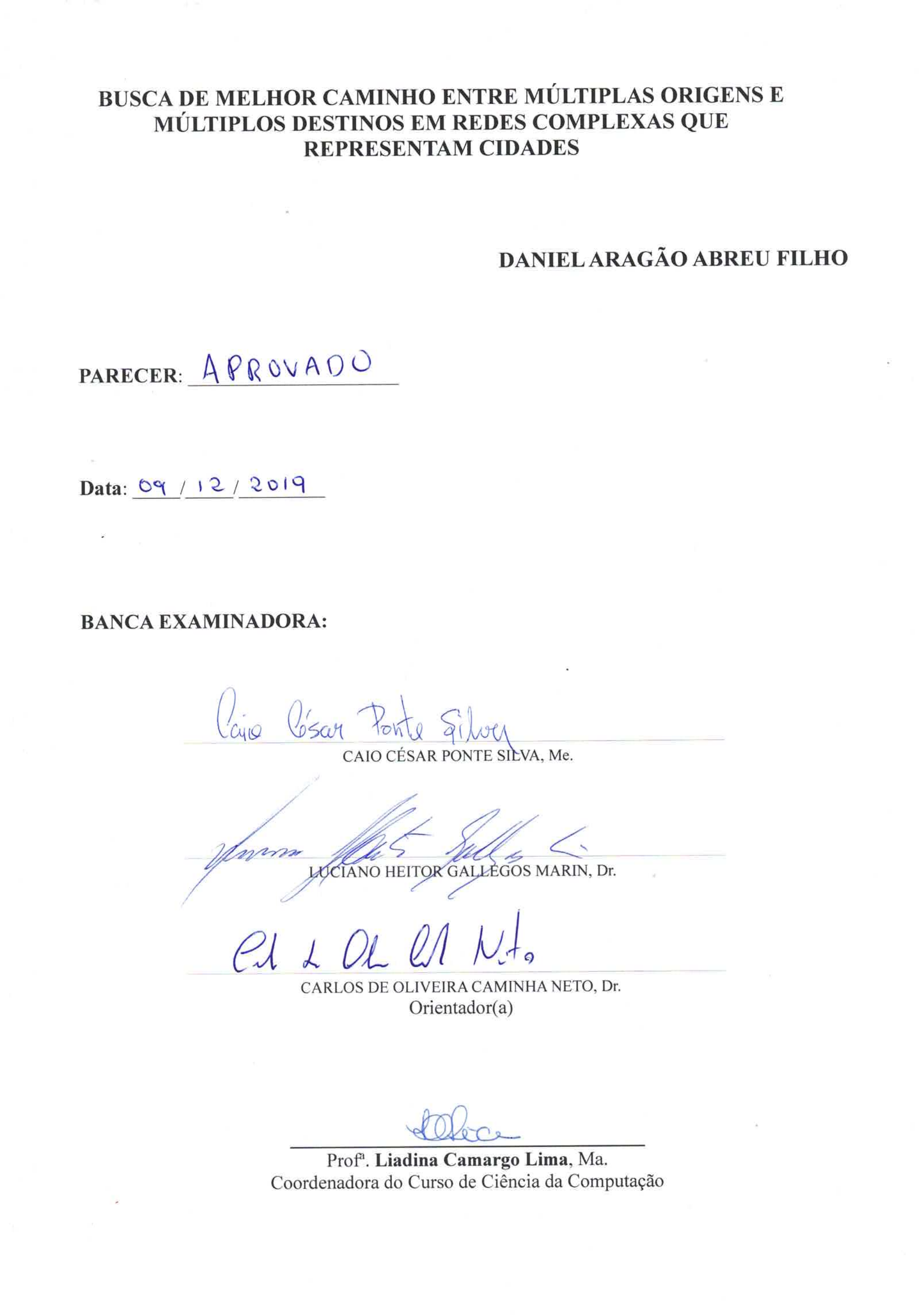}
    %\imprimirfolhadeaprovacao
	
	%\imprimirdedicatoria{elementos-pre-textuais/dedicatoria}
	%\imprimiragradecimentos{elementos-pre-textuais/agradecimentos}
	%\imprimirepigrafe{elementos-pre-textuais/epigrafe}
	 %
	\begin{resumo} %
Foi investigado neste trabalho o uso de uma estratégia de busca em um problema para encontrar o melhor caminho entre múltiplas origens e múltiplos destinos. Nesse tipo de problema deve-se decidir entre várias combinações possíveis qual é a melhor origem e o melhor destino, além do melhor caminho entre esses dois locais. Uma dificuldade notável de se resolver problemas de múltiplas origens e destinos é realizar a busca em tempo reduzido. Esta monografia é uma extensão de uma pesquisa anterior onde o problema descrito foi estudado apenas em uma rede de ônibus da cidade de Fortaleza. Esta extensão consistiu em explorar essa estratégia de busca em grafos que representam a estrutura das vias públicas das cidades de Fortaleza, Mumbai e Tóquio. Ao utilizar essa estratégia juntamente com um algoritmo heurístico, distância de Haversine, foi constatado que é possível diminuir consideravelmente o tempo de busca, porém introduzindo um erro causado pela perda da característica admissível da heurística adotada.

\textbf{Palavras-Chave:} Grafos. Redes Complexas. Algoritmo A*. Algoritmo Floyd-Warshall. %
	\end{resumo} %

	\begin{resumo}[Abstract] %

Was investigated in this paper the use of a search strategy in the problem of finding the best path among multiple origins and multiple destinations. In this kind of problem, it must be decided within a lot of combinations which is the best origin and the best destination, and also the best path between these two regions. One remarkable difficulty to answer this sort of problem is to perform the search in a reduced time. This monography is a extension of previous research in which the problem described here was studied only in a bus network in the city of Fortaleza. This extension consisted of an exploration of the search strategy in graphs that represent public ways in cities like Fortaleza, Mumbai and Tokyo. Using this strategy with a heuristic algorithm, Haversine distance, was noticed that is possible to reduce substantially the time of the search, but introducing an error because of the loss of the admissible characteristic of  the heuristic function applied.

	\end{resumo} %

	\imprimirlistadeilustracoes
	\imprimirlistadetabelas
	\imprimirlistadequadros
	%\imprimirlistadealgoritmos
	%\imprimirlistadecodigosfonte
	\imprimirlistadeabreviaturasesiglas
	%\imprimirlistadesimbolos{elementos-pre-textuais/lista-de-simbolos}
	\imprimirsumario

	%Elementos textuais
	\textual
	\chapter{Introdução}
\label{cap:introducao}
Na literatura é possível encontrar diversas questões relacionadas a estrutura de grafos, dentre elas o problema de busca de melhor caminho, uma classe de problemas bastante conhecida e com muitas aplicações. Algumas variações desse problema são \textit{single source}, na qual tem como objetivo encontrar um caminho de um vértice de origem para todos os outros vértices do grafo, e \textit{point-to-point}, retratado como um problema de encontrar o menor caminho entre dois vértices específicos. Dentre as questões de busca, o chamado de P2P (\textit{Point to Point Shortest Path Problem}) consiste na busca do menor caminho entre dois vértices e pode ser resolvido por vários algoritmos conhecidos, como busca em largura, de custo uniforme, de aprofundamento iterativo e, busca A*, por exemplo \cite{goldberg2005computing, henzinger2016deterministic, bast2016route}.

Dentro dessa classe de problemas P2P, é possível classificar os algoritmos em dois tipos: buscas cegas e buscas com informação. Neste estudo, no entanto, foi baseado na busca com informação ou heurística que por sua vez, utiliza-se de informações prévias sobre o problema, além das suas próprias definições para decidir qual o passo mais útil a ser seguido em determinado instante de uma busca, por exemplo neste trabalho foi utilizada a distância de linha reta na terra, a distância de Haversine, juntamente da posições geográficas informadas, para decidir qual o próximo passo da busca \cite{norvig2004}. Sendo assim, a busca A* é uma das mais conhecidas e que se utiliza de uma função heurística para encontrar resultados ótimos com menores valores de tempo.

\begin{figure}[ht]
 \centering
 \Caption{\label{fig:regioes-onibus} Exemplo de um usuário de ônibus com três possibilidades de paradas possíveis na origem e mais três no destino, de acordo com o quanto ele está disposto a andar.}	
 \UNIFORfig{}{
 	\fbox{\includegraphics[width=16cm]{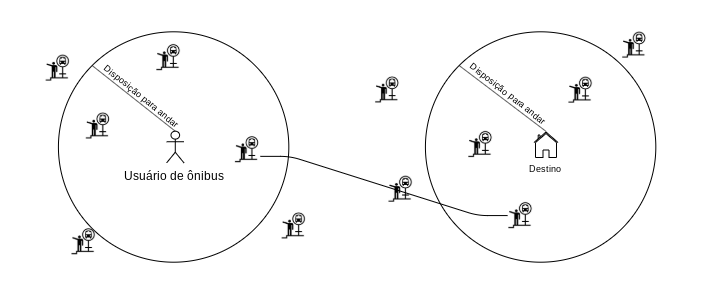}}
 }{
 	\Fonte{Elaborado pelo autor}
 }	
\end{figure}

Os estudos apontam diversas soluções para resolver os problemas P2P, porém o mesmo não pode se afirmar a respeito de problemas de \gls{MOMD}. O \gls{MOMD}, é detalhado como uma busca entre duas áreas de um grafo e não só dois vértices, no qual uma área pode conter um ou mais vértices \cite{bartak2016multiple}. Essa característica pode ser observada n Figura \ref{fig:regioes-onibus}

Para contextualizar o problema de \gls{MOMD}, imagina-se uma cidade onde existe um usuário de ônibus que deseja chegar ao seu destino através de uma linha de ônibus qualquer a partir de sua origem. Porém, esse usuário deve escolher entre diversos pontos de ônibus de acordo com a sua disponibilidade para andar, gerando diversas possibilidades de paradas em que ele pode embarcar. Além disso, o mesmo se aplica ao destino, em que há várias paradas de ônibus nas quais o usuário pode desembarcar de acordo com a sua disponibilidade para andar até o seu objetivo. Esse cenário pode ser observado na Figura \ref{fig:regioes-onibus}.

Compreendendo a situação descrita acima, torna-se relevante afirmar que diversos elementos perpassam essa circunstância, visto que, em uma mesma cidade podem existir pessoas com diferentes posições de origens e destinos, assim como, diferentes disposições para se locomover. Dessa forma, precisa ser delineado diferentes raios para cada individuo, alternando a quantidade de origens e destinos disponíveis em cada busca. 

Um estudo sobre esse tipo de problema \gls{MOMD} para a rede de ônibus da cidade de Fortaleza modelou as paradas de ônibus como vértices e linhas de ônibus como arestas, e utilizou uma estratégia de busca na qual as regiões de vértices de origem e de destino são agrupadas em dois super vértices, para, então, executar o algoritmo de busca A* entre esses dois super vértices \cite{ponte2016busca}. Foi observado que a busca deixa de ser ótima, mas ainda mantém uma acurácia e um desempenho satisfatórios, com baixos valores de erro.

\section{Motivação}
\label{sec:motivacao}

A solução, porém, explora apenas a rede de ônibus da cidade de Fortaleza, criando a necessidade de uma reavaliação da estratégia proposta. Isso pode se tornar possível ao modificar os tipos de grafos utilizados e ainda sim, seria possível evidenciar resultados favoráveis para a estratégia proposta.

A proposta de construção deste trabalho surgiu a partir da demanda de conhecimentos sobre as estratégias do uso de redes  distintas, em diferentes contextos de variabilidade de tamanhos. De forma que, proporcionou um maior conhecimento sobre as propriedades de redes complexas através do uso de redes sintéticas. Dessa forma, busca-se investigar a possibilidade da utilização de um algoritmo de transporte mais rápido que possa ser aplicado em diferentes cenários, promovendo, então, um maior conhecimento sobre essa temática no âmbito científico ao se expandir as possibilidades de aplicabilidade.

\section{Objetivos}
\label{sec:objetivos}

\subsection{Objetivo Geral}
\label{sec:objetivo-geral}

Estender o trabalho de Ponte \textit{et al} (2016), buscando avaliar a utilização dessa estratégia em redes de cidades reais, assim como, propor diferentes cenários para avaliação que podem variar em topologia e tamanho.

\subsection{Objetivos Específicos}
\label{sec:objetivos-especificos}

\begin{itemize}
  \item Verificar se as cidades reais têm topologias similares aos das topologias de redes complexas clássicas ao comparar redes reais e sintéticas.
  \item Compreender como as propriedades de redes complexas podem afetar o modelo proposto por Ponte \textit{et al} (2016) mediante o uso das redes sintéticas.
\end{itemize}

\section{Estrutura do trabalho}
\label{sec:monography-structure}

Esse trabalho é composto por sete partes. A primeira é a Introdução, seguida pela Fundamentação Teórica, a qual é introduzindo conceitos como grafo, buscas e redes complexas. Já a terceira parte, referencia-se ao estado da arte, baseando-se nos trabalhos relacionados, explicando mais detalhadamente sobre a estratégia desenvolvida por Ponte \textit{et al} (2016). O quarto refere-se à metodologia que será adotada nesta pesquisa. Os últimos três capítulos são referentes aos Resultados, Discussão e Conclusão, respectivamente.
	\chapter{Fundamentação Teórica}
\label{cap:fundamentacao-teorica}

\section{Grafos}
\label{sec:graphs-structure}

\begin{figure}[h!]
		\centering
		\Caption{\label{fig:grafo-example} Grafo formado pelo conjunto de Vértices = \{V1, V2, V3, V4, V5\} e arestas = \{E1, E2, E3, E4\}}
		\UNIFORfig{}{
			\fbox{\includegraphics[width=8cm]{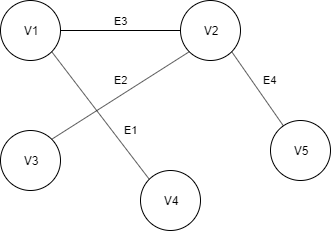}}
		}{
			\Fonte{Elaborado pelo autor}
		}	
	\end{figure}

Os grafos são estruturas de dados essenciais para entender o que é uma rede complexa. Essas estruturas podem ser definidas por \textit{G = (V, E)}, em que \textit{V} é um conjunto de vértices e \textit{E} um conjunto de arestas que ligam esses vértices, ou seja cada \textit{e} $\in$ \textit{E} é um par não ordenado de vértices \textit{\{v, w\}}, em que \textit{v,w} $\in$ \textit{V} \cite{feofiloff2011introduccao}. Isso pode ser observado na estrutura de um grafo na Figura \ref{fig:grafo-example} abaixo.

\begin{figure}[ht]
	\centering
	\Caption{\label{fig:comp-conexo} Grafo com dois componentes conexos \{1, 2, 4, 5\} e \{3, 6\}}	
	\UNIFORfig{}{
		\fbox{\includegraphics[width=8cm]{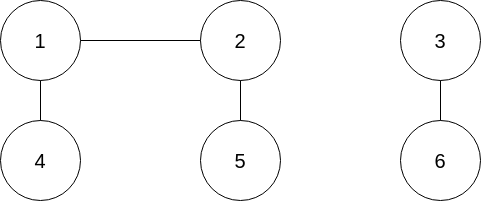}}
	}{
		\Fonte{Elaborado pelo autor}
	}	
\end{figure}

Os grafos são capazes de representar abstrações de ambientes e sistemas reais. Para navegar nessa estrutura, surge o conceito de caminho, este é descrito como a junção de um conjunto de vértices e de arestas que une o ponto de origem ao de destino \cite{newman2010}. Porém, em um grafo nem sempre existe um caminho entre dois vértices, como pode ser observado na Figura \ref{fig:comp-conexo}, visto que não é possível traçar um caminho do vértice 1 para o vértice 6.

Quando não há caminho entre dois vértices, destaca-se que o vértice de origem e de destino estão em componentes conexos distintos, é definido como um subgrafo do grafo original que possui caminhos que interliguem todos os vértices. Dessa forma, em um grafo não dirigido para todo vértice A de um componente conexo e para todo vértice B do mesmo componente, sempre haverá um caminho entre A e B ou B e A \cite{chrispino2013area}. Ao observar na Figura \ref{fig:comp-conexo} os componentes distintos, evidencia-se que é impossível chegar no vértice 3 partindo-se do vértice 4, ou seja, eles pertencem a diferentes componentes conexos. Ademais, vale destacar o componente gigante de uma rede como o que possui o maior número de vértices \cite{chrispino2013area}, este componente é uma estrutura composta por um número de vértices que aumenta proporcionalmente ao crescimento do número de vértices do grafo e, como também, contém quase todos os vértices do grafo \cite{newman2010}.

\section{Redes complexas}
\label{sec:complex-networks}

As redes complexas são estruturas que podem exprimir comportamentos. Isso pode ser exemplificado através de uma rede que representa o sistema nervoso, em que os vértices são as células nervosas que, ao mesmo tempo, transmitem sensações em escalas macroscópicas, e sinapses nervosas em escalas microscópicas \cite{sayama2015}. Essas redes apresentam classificações que serão posteriormente contextualizadas no Tópico \ref{sec:topologias-de-redes}.

Ainda sobre esse tipo de estrutura de redes complexas, pode-se destacar outros exemplos, como as redes de cadeias alimentares ou a internet. Estas são redes que possuem numerosos componentes interagindo entre si e com a sua própria organização, como também, evoluem no tempo em tamanho e em forma podendo gerar novos comportamentos. Além disso, por muito tempo supôs-se que eram sistemas aleatórios, entretanto, são regidos por complexas formas de organização \cite{sayama2015, albert2002statistical}.

\subsection{Propriedades}
\label{sec:fundamentacao-teorica-propriedades}

Algumas propriedades são usadas comumente para caracterizar essas redes complexas. O primeiro conceito da lista é a entropia de \textit{shannon}, advinda da área de teoria da informação, uma medida que traz uma taxa de surpresa \cite{vajapeyam2014}. A entropia é um número real positivo utilizado para tentar prever qual o valor de uma variável conhecida baseada no histórico de valores já gerados. Deste modo, é possível informar quão aleatória é a estrutura de um grafo baseando-se na distribuição dos possíveis valores de graus dos vértices do grafo \cite{lewis.c2.a2009}, sendo o grau o número de arestas que incidem em um vértice \cite{oliveira2012monitoramento}. Uma outra propriedade relevante é o conceito de \textit{hub}, conceito que representa os vértices com grande quantidade de arestas \cite{lewis.c2.a2009} conectadas a eles.

Entre as principais características de uma rede, têm-se o comprimento de caminho médio, definido pela média de arestas do menor caminho entre cada par de vértices de um grafo, ou seja, é a média da quantidade de saltos que são necessários para se alcançar um vértice qualquer a partir de um outro vértice qualquer da mesma rede \cite{wang2003complex, lewis.c2.a2009}.

Sobre a última propriedade, o coeficiente de \textit{cluster} mede quão aglomerado é o grafo por meio do cálculo da média de C para cada vértice da rede: $C = 2 \cdot n/k\cdot(k - 1)$, sendo \textit{k} a quantidade de vizinhos do vértice e \textit{n} o número de arestas entre os \textit{k} vizinhos do vértice \cite{ravasz2003hierarchical,wang2003complex,newman2010}. Portanto, é usado que para definir o quanto os vértices estão conectados.

%Aproveitando o mesmo exemplo de cidades, é notável que muitas vezes ao dia as pessoas precisem trafegar entre os vértices e arestas de suas cidades, sendo assim, se uma mulher deseja sair de um ponto A e chegar a um ponto B de uma cidade ela irá precisar traçar um caminho que no grafo representa um conceito definido por um conjunto de vértices e arestas que de forma ordenada conectam vértices do grafo  \cite{newman2010}. Porém é necessário entender que nem sempre haverá caminhos que conectem esses dois pontos, mostrando que o grafo não é formado de um componente só, observar a figura \ref{fig:comp-conexo}.

%Diametro
%A graph’s diameter,
%d.G/, is the longest path between any two
%nodes in the graph
% \cite{bunn2000}

%Cluster Coefficient
%Hub degree
%Avg. Path. Length.

\subsection{Topologias de redes/Tipos de redes complexas}
\label{sec:topologias-de-redes}

\begin{figure}[ht]
		\centering
		\Caption{\label{fig:regular-example} Exemplo de grafo regular com entropia próxima ou igual a zero. Obs.: as arestas das bordas estão sendo omitidas por questão de visualização.}
		\UNIFORfig{}{
			\fbox{\includegraphics[width=8cm]{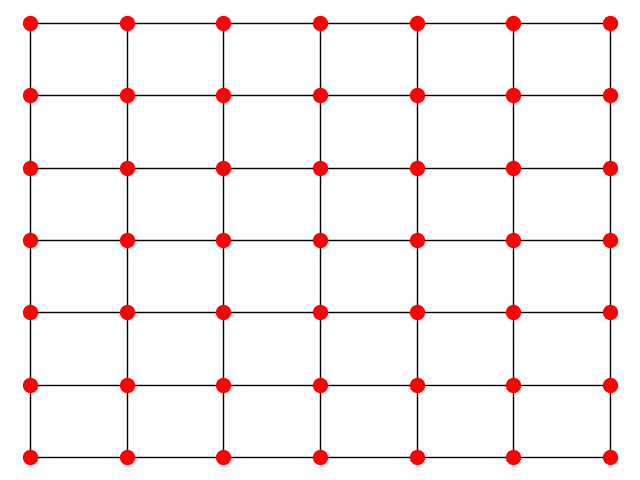}}
		}{
			\Fonte{Elaborado pelo autor}
		}	
	\end{figure}

\begin{figure}[ht]
		\centering
		\Caption{\label{fig:random-example} Rede aleatória com entropia próxima ou igual a um. Obs.: as arestas das bordas estão sendo omitidas por questão de visualização.}
		\UNIFORfig{}{
			\fbox{\includegraphics[width=8cm]{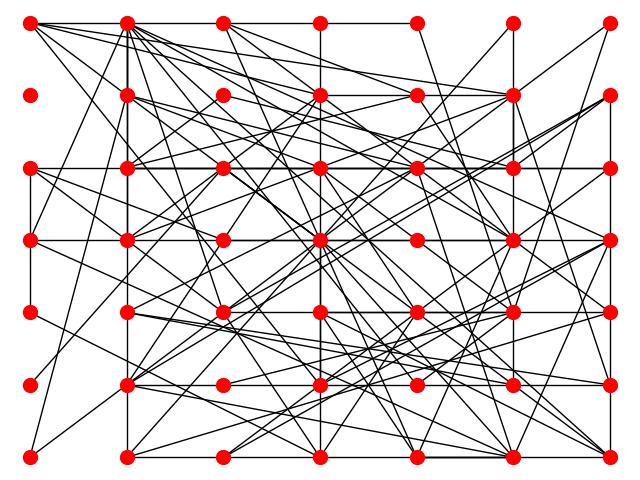}}
		}{
			\Fonte{Elaborado pelo autor}
		}	
	\end{figure}

As redes complexas podem ser classificadas em quatro topologias básicas: regular, aleatória, mundo pequeno e livre de escala. O formato mais simples seria a de um grafo regular que possui um padrão de distribuição de suas arestas. Na mesma quantidade para cada vértice, ou seja, uma entropia igual a zero, já que na entropia de \textit{shannon} possui o 1 como o grau de aleatoriedade máxima \cite{lewis.c2.a2009}. Na Figura \ref{fig:regular-example}, todos os vértices têm o mesmo número de arestas, representando o grafo regular. Ao contrário dele tem-se o grafo aleatório, entropia próxima de 1, onde para cada vértice no grafo há uma possibilidade de conexão com outro vértice, perdendo o padrão de distribuição apresentado pelo regular, como mostra a Figura \ref{fig:random-example} \cite{lewis.c2.a2009, antiqueira2005modelando}.

\begin{figure}[ht]
		\centering
		\Caption{\label{fig:sw01-example} Exemplo de grafo mundo pequeno com \textit{p = 0,1}. Obs.: as arestas das bordas estão sendo omitidas por questão de visualização.}
		\UNIFORfig{}{
			\fbox{\includegraphics[width=8cm]{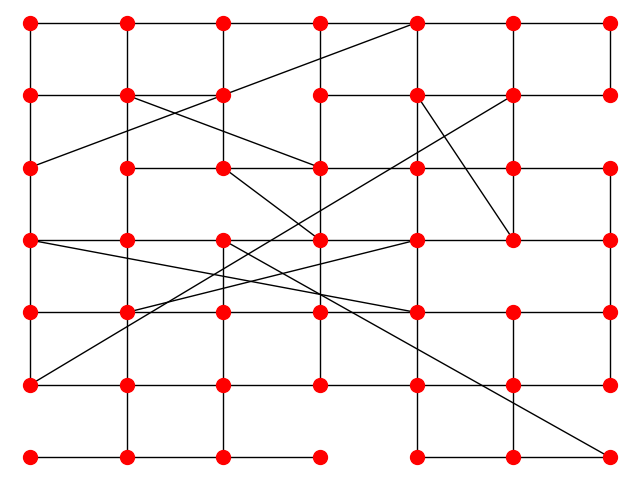}}
		}{
			\Fonte{Elaborado pelo autor}
		}	
	\end{figure}

\begin{figure}[ht]
		\centering
		\Caption{\label{fig:sw05-example} Exemplo de grafo mundo pequeno com \textit{p = 0,5}. Obs.: as arestas das bordas estão sendo omitidas por questão de visualização.}
		\UNIFORfig{}{
			\fbox{\includegraphics[width=8cm]{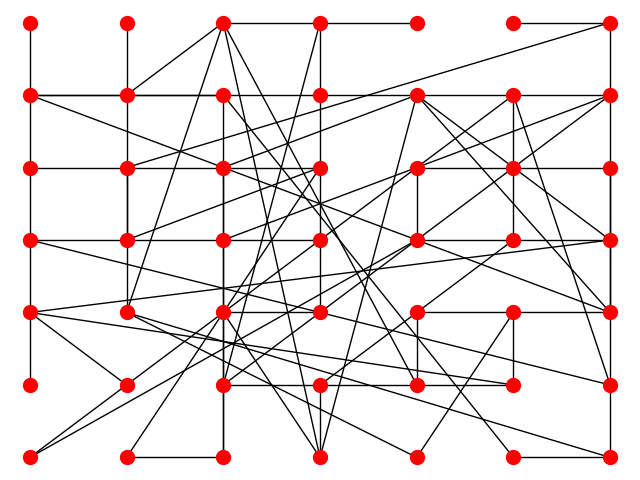}}
		}{
			\Fonte{Elaborado pelo autor}
		}	
	\end{figure}

A próxima topologia que será abordada é conhecida como mundo pequeno. Essas redes tem uma entropia intermediária, entre o regular e o aleatório, pois são a princípio regulares até que suas arestas sejam redistribuídas  de forma aleatória, gerando o crescimento na entropia. Devido a redistribuição, vértices que se encontravam distantes agora podem ser ligados tornando possível realizar um trajeto no grafo por menores caminhos, diminuindo o diâmetro e o coeficiente de \textit{cluster} do grafo. Sendo assim, observa-se um menor comprimento de caminho médio provocado pelo efeito mundo pequeno \cite{wang2003complex, lewis.c2.a2009, caminha2012modeling}. A seguir um exemplo com dois diferentes percentuais \textit{p} de redistribuição de arestas nas Figuras \ref{fig:sw01-example} e \ref{fig:sw05-example}.

\begin{figure}[ht]
		\centering
		\Caption{\label{fig:scalefree-example} Exemplo de grafo livre de escala. Obs.: as arestas das bordas estão sendo omitidas por questão de visualização.}
		\UNIFORfig{}{
			\fbox{\includegraphics[width=8cm]{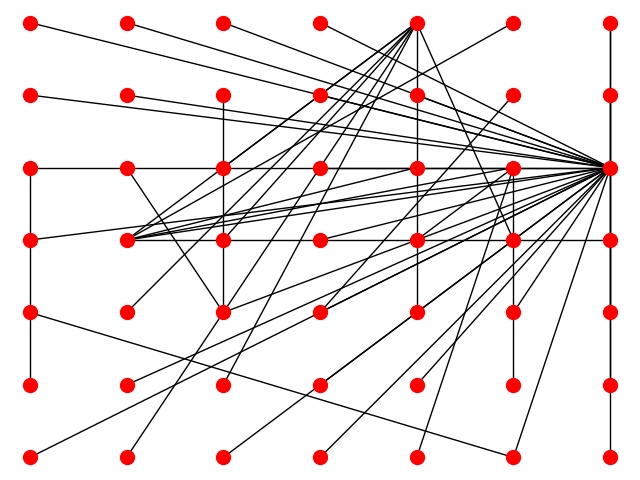}}
		}{
			\Fonte{Elaborado pelo autor}
		}	
	\end{figure}

A próxima topologia é a rede livre de escala. Esta é um tipo de grafo que segue uma lei de potência e obedece duas regras de formação: a regra de crescimento, que consiste em uma rede que inicia-se com poucos vértices em que a cada espaço de tempo pode surgir um vértice novo; e a regra de \textit{preferential attachment}, que ao adicionar um vértice novo, este vai se ligar a um vértice já existente de forma que o vértice escolhido obedeça uma probabilidade proporcional ao seu grau. Isso gera uma rede em que alguns vértices possuem alta concentração de arestas, enquanto outros só possuem uma, como na Figura \ref{fig:scalefree-example}. Dessa forma, caminhar nesse tipo de grafo é mais acessível, pois ao chegar em um vértice conectado há maiores chances de encontrar o vértice destino. Essa topologia também pode representar alguns tipos de redes, como a internet, na qual a distribuição de probabilidade não é mais uniforme como no aleatório e apresenta um crescimento constante \cite{wang2003complex, lewis.c2.a2009}.

\begin{figure}[ht]
    \centering
    \Caption{\label{fig:kiviat-graphic} Propriedade dos grafos \textit{Ring} (regular), \textit{Random} (Aleatório), \textit{Small World} (Mundo pequeno), \textit{Scale free} (Livre de escala).}
    \UNIFORfig{}{
    	\fbox{\includegraphics[width=14cm]{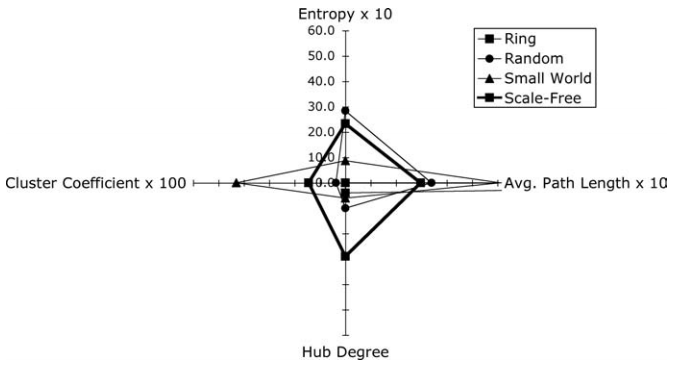}}
    }{
    	\Fonte{\cite{lewis.c6.a2009}}
    }	
\end{figure}

Para apresentar as propriedades em cada uma das topologias, pode ser observado o Quadro \ref{tab:propriedades-vs-topologias} abaixo que mostra cada uma em uma escala de três níveis, baixa, média e alta, com base no gráfico de radar na Figura \ref{fig:kiviat-graphic}.

\begin{quadro}[ht]
\centering
\caption{Comparação entre as topologias com base nas propriedades apresentadas.}
\label{tab:propriedades-vs-topologias}
\begin{tabular}{|c|c|c|c|c|}
\hline
Propriedade                  & Regular & Aleatória & Mundo pequeno & Livre de escala \\ \hline
Entropia                        & Baixo   & Alta      & Média       & Alta       \\ \hline
\textit{Hub}                    & Baixo   & Baixo     & Baixo       & Alto       \\ \hline
Comp. de caminho médio          & Baixo   & Médio     & Alto        & Médio      \\ \hline
Coeficiente de \textit{cluster} & Baixo   & Baixo     & Alto        & Médio      \\ \hline
\end{tabular}
\end{quadro}

\section{Caminhos e buscas}
\label{sec:caminhos}

\begin{figure}[ht]
		\centering
		\Caption{\label{fig:path-example} Exemplo de caminho.}
		\UNIFORfig{}{
			\fbox{\includegraphics[width=8cm]{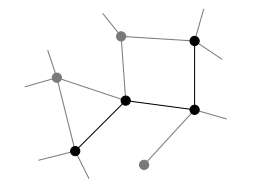}}
		}{
			\Fonte{\cite{newman2010}}
		}	
	\end{figure}

Para compreender melhor o resultado de um algoritmo de busca, define-se caminho como qualquer sequência de vértices, de forma que cada par da sequência seja ligado por uma aresta e que não haja repetição de nenhum vértice, conforme ilustrado na Figura \ref{fig:path-example} \cite{newman2010}. Quando obtém-se o melhor caminho possível, este é chamado de caminho ótimo.

\subsection{Algoritmos de Buscas}
\label{sec:algoritmo-de-busca}

Diversos tipos de problemas podem ser descritos com busca em grafos, por exemplo, como o roteamento do tráfego de rede telefônica, o mapeamento das relações interpessoais em redes sociais ou, no contexto desse trabalho, a busca para encontrar o melhor caminho em uma cidade. Devido a variedade de problemas de busca, é necessário o uso de algoritmos que sejam capazes de encontrar uma solução ótima para o problema em tempo viável \cite{hart1968formal}. Portanto, utilizou-se o algoritmo A* que é um tipo de busca informada, ou busca heurística com desempenho e soluções satisfatórios.

\subsubsection{Algioritmo Heurístico}
\label{sec:heuristicas}

\begin{figure}[ht]
		\centering
		\Caption{\label{fig:desigualdade-triangulos} Demonstração da propriedade da consistência em uma heurística definida como uma função de distância euclidiana.}
		\UNIFORfig{}{
			\fbox{\includegraphics[width=10cm]{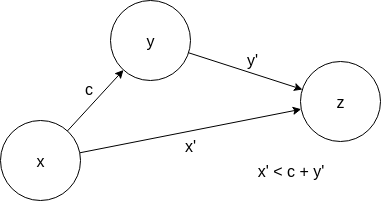}}
		}{
			\Fonte{Elaborado pelo autor}
		}	
	\end{figure}

Compreendendo um algoritmo heurístico, definidas estas como funções que resultam em estimativas de uma solução real. No caso deste trabalho são estimativas que tentam calcular o valor real de uma medida de distância. Por exemplo, quando se precisa do menor caminho entre uma loja e outra dentro de um shopping, provavelmente o caminho não será uma linha reta. Pode-se realizar essa inferência por haver vários obstáculos, como paredes, pessoas e outros objetos, porém é possível estimar qual a distância real a partir da heurística que informa a distância em linha reta. Se essa solução avaliada for sempre menor que a solução real para determinado problema, têm-se uma heurística admissível. Outra propriedade da heurística é a consistência que remete à regra de desigualdade de triângulos, demonstrado na figura \ref{fig:desigualdade-triangulos}, esta afirma que ao sair de um ponto \textit{x} e chegar em outro \textit{y}, a estimativa \textit{x'} de solução a partir de \textit{x} sempre será menor que o custo \textit{c} de se chegar em \textit{y} partindo de \textit{x} somado a estimativa \textit{y'} observada de \textit{y} \cite{norvig2004}.

\subsubsection{Busca A*}
\label{sec:astart}

Após entender algoritmos heurísticos, a busca heurística é aquela que utiliza de informações além das definições do problema para decidir qual o passo mais útil a ser seguido em determinado instante de uma busca \cite{norvig2004}, ou seja, ela é atribuída a uma heurística. Então, utilizando a busca A*, é possível encontrar soluções ótimas ao se utilizar de uma função heurística admissível e consistente. A função heurística utilizada para o problema deste trabalho foi o cálculo da distância em linha reta com uma função euclidiana utilizada para um plano cartesiano, e a fórmula da distância de Haversine que calcula algo semelhante a linha reta em objetos esféricos como a terra.

\begin{figure}[ht]
		\centering
		\Caption{\label{fig:exemplo-heuristica} Tendo a origem como o vértice v1 e o destino o vértice v5, durante a busca com o vértice v3 selecionado temos o valor em destaque de \textit{g}(v3) e de \textit{h}(v3) totalizando \textit{f}(v3) = 12}
		\UNIFORfig{}{
			\fbox{\includegraphics[width=10cm]{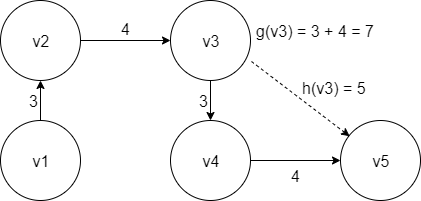}}
		}{
			\Fonte{Elaborado pelo autor}
		}	
	\end{figure}

Retomando a busca A*, o algoritmo avalia suas decisões por meio de uma função de avaliação $f(n) = g(n) + h(n)$, em que \textit{n} representa um vértice e a função representa o custo de se escolher aquele vértice \textit{n} como uma parte da solução para o caminho ótimo. Esse \textit{f(n)} pode ser decomposto em duas novas funções, \textit{g(n)} que representa o custo atual de se atingir aquele vértice \textit{n} a partir da origem e \textit{h(n)} que é a própria função heurística de se chegar ao vértice objetivo a partir do vértice \textit{n} \cite{hart1968formal}. Um exemplo que ilustra a função de avaliação pode ser observado na Figura \ref{fig:exemplo-heuristica}.

O algoritmo funciona de forma que a partir do vértice inicial, se calcula o \textit{f(n)} de cada um dos seus vértices vizinhos e adiciona-se seus resultados numa lista chamada de fronteira. Dessa lista, é retirado o vértice com o menor valor de \textit{f(n)} selecionando-o para continuar a busca até que o vértice escolhido seja o vértice objetivo. Isso faz com que a busca sempre tenda a andar na direção do vértice objetivo, diminuindo a seleção, expansão, de vértices que não serão parte do caminho ótimo para o problema \cite{norvig2004,hart1968formal}.

\subsection{Colapsar ou \textit{Coarsening}}
\label{sec:coarsening}

O processo de colapsar é tal que, dado uma região de um grafo, consiste em transformar todos os vértices dessa região em um único vértice, onde as arestas incidentes a esse vértice são as arestas que se ligam com os vértices que não estão na região, quanto as arestas incidentes a vértices dentro da região, estas são colapsadas também
\cite{karypis1995metis}. A posição do vértice formado assumirá a posição do vértice mais central dada a região escolhida.

\begin{figure}[ht]
    \centering
    \Caption{\label{fig:colapsar-vertices} Processo de colapsar vértices.}	
    \UNIFORfig{}{
     \fbox{\includegraphics[width=10cm]{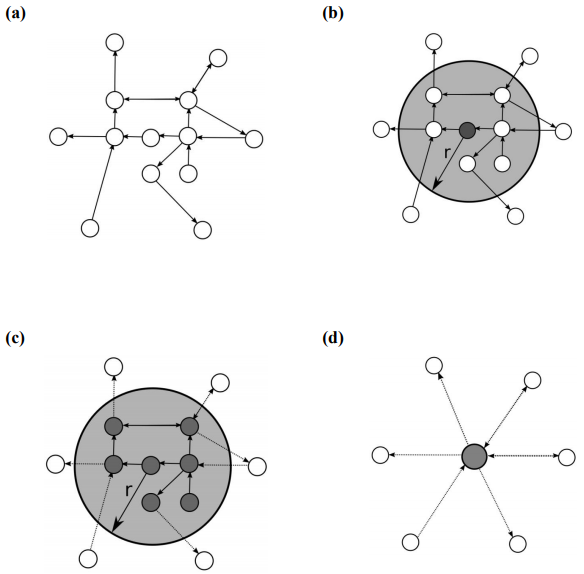}}
    }{
     \Fonte{\cite{ponte2016busca}}
    }
 \end{figure}

Um exemplo mais preciso do que consiste o colapso está na Figura \ref{fig:colapsar-vertices}. No item \ref{fig:colapsar-vertices}(a) é possível notar um grafo. Na Figura \ref{fig:colapsar-vertices}(b), é possível observar uma região selecionada para colapsar. Essa região é demarcada a partir da disponibilidade do usuário de ônibus para caminhar numa distância \textit{r}. Na Figura \ref{fig:colapsar-vertices}(c), são destacados os vértices que serão colapsados em um único super vértice, de modo que cada vértice represente uma parada de ônibus. Por fim, na Figura \ref{fig:colapsar-vertices}(d), o resultado de colapsar, quando um super vértice está presente no lugar de todos os vértices da Sub Figura (c).

Ao encontrar um caminho após uma busca entre nós colapsados, tem-se uma aresta final, a qual foi utilizada para encontrar o super vértice de destino. Para resgatar qual o vértice destino real dentro da região colapsada, basta olhar em qual vértice incidia aquela aresta final antes de colapsar a região. Sendo assim o processo inverso de colapsar acontece para recuperar qual o verdadeiro destino.
 
\subsubsection{Heurística não admissível}
\label{sec:heuristicas-n-admissivel}

\begin{figure}[ht]
    \centering
    \Caption{\label{fig:colapsar-n-admissivel} Uma busca que encontra uma heurística maior que o resultado final escolhido.}	
    \UNIFORfig{}{
     \fbox{\includegraphics[width=12cm]{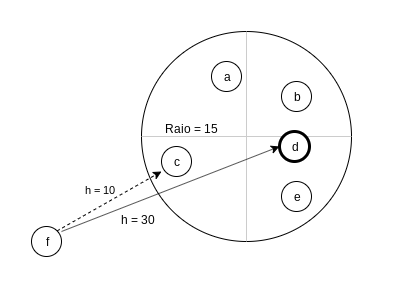}}
    }{
     \Fonte{Elaborado pelo autor}
    }
 \end{figure}

Ao colapsar os vértices de uma região e tomando-se como posição os valores do vértice mais central, o destino final não necessariamente será esse vértice. Desse comportamento observa-se um erro que pode variar de acordo com o raio da região, pois a posição do super vértice pode ser pior que a posição do vértice destino, sendo assim a heurística da linha reta não informará um resultado menor ou igual ao valor real, e sim, um valor que pode ser menor ou maior excedendo-se no máximo a um valor próximo ao diâmetro da região colapsada. Um exemplo pode ser visto na Figura \ref{fig:colapsar-n-admissivel}, onde o raio com tamanho 15 pode proporcionar um erro de no máximo 30, caso o vértice selecionado como central esteja numa extremidade e o vértice objetivo esteja na outra extremidade da região. A seta pontilhada demonstra a heurística correta e representa o vértice final. A seta sólida representa a heurística calculada, porém não é o vértice final. O vértice \textit{f} representa a atual posição de busca e o \textit{d} o vértice escolhido como mais central.

\section{Algoritmo Floyd-Warshall}
\label{sec:floyd-warshall}
Uma forma de se descobrir todos os caminhos possíveis de um grafo é utilizando o algoritmo de Floyd-Warshall que é de complexidade \textit{O($V^3$)} onde \textit{V} é número de vértices do grafo, por consequência os caminhos possíveis entre duas regiões específicas seria apenas iterar sobre o resultado. Para executar tal algoritmo é necessário uma estrutura de repetição com três laços aninhados e uma estrutura de dados de matriz  \textit{V x V} para armazenar os resultados \cite{gaioso2013paralelizaccao}. Apesar da existência desse algoritmo, ele não pode ser utilizado por tornaria o problema muito mais complexo, já que itera sobre todos os nós do grafo, não somente duas regiões específicas. Isso é um comportamento que não pode se evitar, pois não seria possível reduzir a busca já que não se sabe quais os vértices que serão utilizados antes do tempo de execução.
	\chapter{Trabalhos Relacionados}
\label{cap:trabalhos-relacionados}

A busca de um melhor caminho entre \gls{MOMD} é um tema discutido no trabalho de Ponte \textit{et al} (2016). No trabalho, foi abordada uma estratégia para realizar buscas em redes de ônibus. A partir da disponibilidade do usuário de ônibus de caminhar, o número de paradas de ônibus poderia variar na origem e no destino, de forma que a busca tenha que ser executada entre duas áreas. O processo realizado foi de colapsar os vértices de origem possíveis em um único super vértice, fazendo o mesmo no destino, logo a busca executada fora somente entre dois vértices. Foi mostrado que a estratégia executada foi satisfatória quanto ao tempo e a acurácia, ou seja, apesar do resultado não ser ótimo, o algoritmo continua satisfatório.

O trabalho de Ponte \textit{et al} (2016) comparou sua estratégia com um método que foi chamado de Força Bruta. Nesse método são feitas $n^2$ buscas, para \textit{n} igual ao número de vértices na origem e no destino, onde cada nó de origem é testado com cada nó de destino. Isso não significa que não poderiam ser dois números distintos, porém foram assumidos origem e destino como \textit{n}. No trabalho realizado pelos referidos autores, eles obtiveram como resposta que colapsar os vértices produz resultados com pouca variação de tempo de execução, mesmo aumentando o raio de disposição do usuário de transporte, ou seja, com o aumento de vértices nas regiões, não há diferença significativa no tempo de execução. Porém, ao utilizar o método de Força Bruta como comparativo à sua pesquisa, eles observaram que o desempenho foi bastante reduzido com o aumento do raio, resultando em um aumento significativo no tempo de execução para poucas alterações incrementais no raio.

A Acurácia empregada pelos autores mencionados logo acima foi outro resultado importante para a realização deste projeto de pesquisa. A partir dessa métrica, foi criada uma medida para destacar quantos caminhos eram ótimos ao utilizar a estratégia de colapsar. Para isto, foi necessário executar o método de Força Bruta, citado anteriormente. Como resposta observou-se que ao aumentar o raio a acurácia caiu. Entretanto também foi possível observar que o erro médio das rotas não ótimas não foi muito significativo, ou seja, apesar de não ser mais um caminho ótimo possui resultado próximo do ótimo \cite{ponte2016busca}.

O problema de \gls{MOMD} também foi solucionado por meio de otimização em outro trabalho. A função objetivo é uma minimização do custo total de arestas. A motivação para este trabalho também foi de uso de redes de transporte e foi provado ser um problema NP-Completo que pode ser fundamentado pelo modelo SAT, baseado-se em preservação de fluxo. Foi observado que não haviam modelos com tempo viável até o determinado momento deste trabalho de otimização \cite{bartak2016multiple}.

Além disso, encontrar bons caminhos de forma eficiente podem também afetar positivamente a redução da criminalidade, visto que o fluxo de pessoas nas cidades influenciam a incidência de crimes. Dessa forma, entender esse tipo de fluxo pode influenciar no planejamento de medidas interventivas de segurança pública, permitindo um melhor direcionamento do policiamento para a prevenção e remediação de crimes \cite{caminha2017human}.

Outro estudos também destacam que o melhor entendimento dessas redes complexas de cidades podem aprimorar os fluxos da rede de transporte público, como retirar gargalos ou sugerir novas rotas mais rápidas ou menores \cite{caminha2016micro}. Esse mesmo estudo também propôs maneiras de se coletar informações de origem e destino a partir de onde o usuário de ônibus passa seu vale transporte.

Apesar de muitas vezes procurar o melhor caminho, nem sempre esse é o objetivo do usuário de ônibus. Na verdade a maioria dos usuários preferem tomar caminhos mais seguros que caminhos mais rápidos passando pelo terminal. Isso se deve principalmente que 98\% das linhas de ônibus passam por regiões com mais de dez crimes \cite{sullivan2017towards}.
	\chapter{Metodologia e conjunto de dados}
\label{chap:metodologia}

Para realizar o trabalho, foi explorado um conjuntos de dados de grafos que representam cidades reais. Os dados das vias da cidade foram coletados a partir do software de mapeamento de código aberto \textit{OpenStreetMap}. Utilizando sua \textit{\gls{API}} online, foi feito o \textit{download} de grafos de cidades em formato de arquivo \textit{\gls{XML}} e convertidos em uma arquivo com estrutura própria para diminuir o consumo de dados e facilitar a leitura. Nesses arquivos são armazenadas informações como ruas, nomes de ruas, cruzamentos georreferenciados e identificadores. %\textit{Extensible Markup Language} (XML) 

A partir desses arquivos, foram montados os grafos para cada cidade, onde cada vértice do grafo representa uma esquina ou curva. Já as arestas são as ruas entre cada esquina, formando assim um grafo semi-planar com uma média, entre as três cidades, de 62 mil vértices e 110 mil arestas. Observar Tabela \ref{tab:city-verticies} com os dados de vértices e arestas para cada cidade. %Os atributos das arestas são nomes das ruas e tamanho, além dos vértices de origem e destino, já os vértices possuem coordenadas para georreferenciamento e um identificador.

\begin{table}[h]
\centering
\caption{Número de vértices e arestas para cada grafo de cidade baixada}
\label{tab:city-verticies}
\begin{tabular}{c c c}
\hline
Propriedade & Número de vértices & Número de arestas \\ \hline
Mumbai      & 24388              & 43983             \\ 
Fortaleza   & 47138              & 104658            \\ 
%Paris       & 50720              & 77791             \\ 
Tóquio      & 116057             & 183573            \\ 
%Nova Iorque & 116661             & 329480            \\ \hline
\hline
\end{tabular}
\end{table}

Em paralelo sobre o segundo conjunto de dados, foi desenvolvido um gerador de grafos que obedece a mesma estrutura de arquivos de armazenamento das cidades. Dessa forma foram gerados de grafos em quatro topologias diferentes. As topologias utilizadas foram aleatório, regular, \textit{small world} (com dois graus de aleatoriedade diferentes) e livre de escala. Em todos os casos foram gerados grafos com distribuição espacial em formato de grade com tamanho de 10 mil vértices. Os algoritmos de geração foram baseados no trabalho de Lewis (2009).

\begin{figure}[h]
    \centering
    \Caption{\label{fig:mapa-fortaleza} Mapas das cidades representados em vértices, onde os eixos são as latitudes e longitudes das cidades.}	
    \UNIFORfig{}{
     \fbox{\includegraphics[width=16cm]{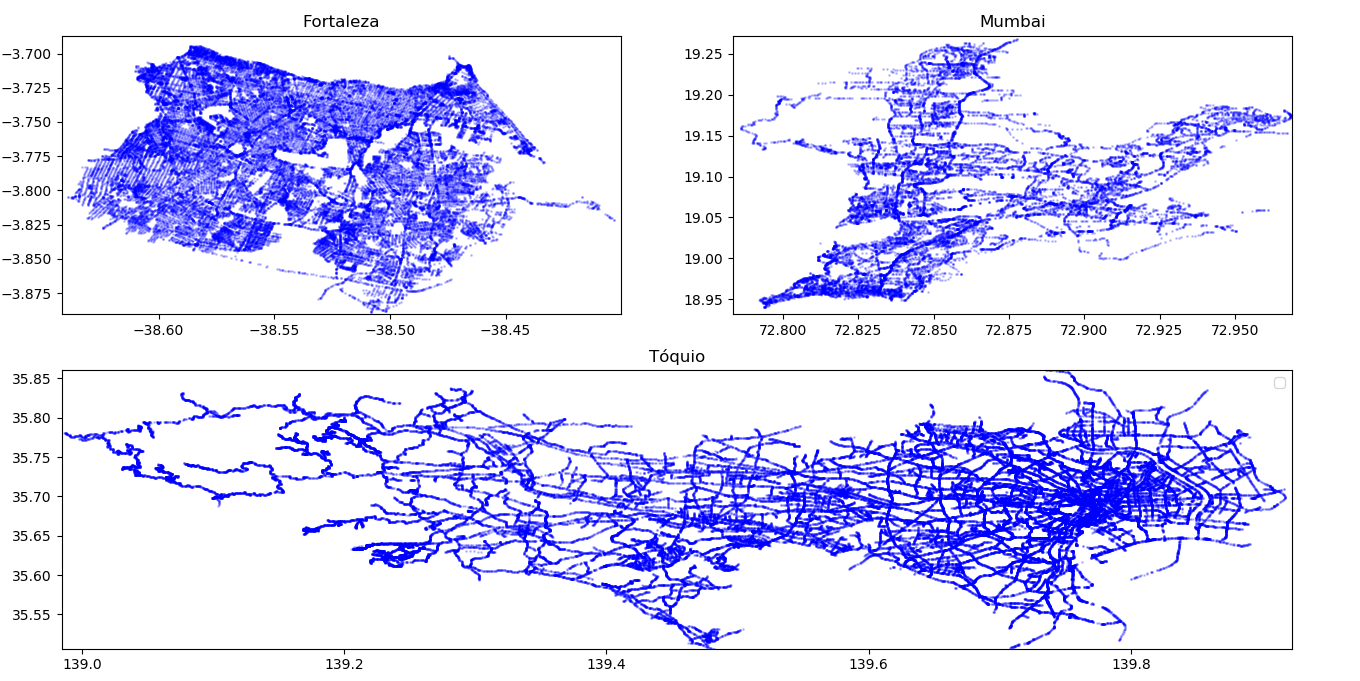}}
    }{
     \Fonte{Elaborado pelo autor}
    }
\end{figure}

Neste contexto, foi necessário ainda realizar uma limpeza nos grafos das cidades baixadas, pois nem sempre o grafo representava o seu maior componente conexo ou apresentava vértices em alguma posição inválida, dessa forma evitando que algumas buscas pudessem encontrar o devido caminho. A limpeza consistiu em executar uma busca pelo maior componente conexo de cada grafo e salvá-los em arquivos distintos com o mesmo formato de leitura. De cada novo arquivo foram selecionados aleatoriamente dez mil pares de vértices de \gls{OD} para cada grafo e armazenados em um novo arquivo. Vale ressaltar que cada par representa uma busca dentro do grafo em questão e portanto uma origem para um indivíduo em um ponto qualquer da cidade, representado por um vértice, com seu destino em outro ponto qualquer da cidade, representado por outro vértice. A Figura \ref{fig:mapa-fortaleza} ilustra os grafos que representam a estrutura das vias das cidades estudadas.

Dado os grafos das cidades, foi possível retirar algumas propriedades básicas a respeito de cada uma e suas características como redes complexas e compará-las a redes sintéticas geradas com as topologias apresentadas no tópico \ref{sec:topologias-de-redes} com a quantidade de 10.000 vértices em que os vértices se distribuem em formato de grade. Abordando a entropia, foi possível observar que as cidades se aproximam a uma topologia de rede de mundo pequeno que são redes semelhantes a redes regulares, porém com algum grau de aleatoriedade. Ao se analisar a quantidade a propriedade de \textit{Hub}, as cidades se mantiveram mais uma vez próximas as redes de mundo pequeno e também de redes regulares. Por último foi notado que a respeito do coeficiente de \textit{cluster}, os grafos das cidades se localizam entre as os grafos do tipo mundo pequeno mais uma vez e aleatório, simbolizando uma grande similaridade com as redes de mundo pequeno. A propriedade de Comprimento de caminho mínimo não pode ser analisada devido ao seu custo de execução que foi limitado pelo tempo.

% fzr um paragrafo explicando fb e colapsar
% revisar
%Quanto às buscas, cada arquivo de \gls{OD} gerado será submetido a dez configurações de execução, formadas por uma variação entre os cinco raios para cada uma das duas estratégias de busca. Isso acontece porque o raio simula o quanto o usuário de ônibus está disposto a andar, considerando também diferentes perfis de pessoas dentro de uma cidade, podendo variar por exemplo de acordo com sua faixa etária. Um adolescente e um idoso possuem diferentes disposições de movimentação. Ainda sobre o parâmetro de raio, tem-se que os vértices não são mais paradas de ônibus, mas cruzamentos e curvas. Para uma busca de tamanho 50 metros significa dizer que serão parte do conjunto de origem todos os cruzamentos e curvas dentro do raio de cinquenta metros e o mesmo vale para o conjunto de vértices de destino. A segunda configuração seria a de estratégia que varia apenas em Colapsar e Força Bruta. Para um terceiro parâmetro temos as três cidades disponíveis em seus respectivos grafos.

A estratégia de Força Bruta foi utilizada como uma variante do algoritmo de \textit{Floyd-Warshall}, compreendendo que a tabela com os registros dos melhores caminhos de busca é construída somente com as regiões de origem e destino. Apesar disso, o algoritmo continua com a complexidade de tempo cúbica, mostrando-se ineficiente. Entretanto, a estratégia de Força Bruta foi utilizado para efeito de comparação, permitindo, então, apontar os resultados que fossem ótimos e o ganho de tempo da estratégia de Colapsar.
Além disso, é importante salientar que a Força Bruta representa uma busca entre todas as origens em direção a todos os destinos, de forma que seja realizado uma quantidade de buscas em função do número de vértices na origem e no destino. Sobre o Colapsar, este permite que ao realizar a unificação de todos os vértices de origem em uma única origem, assim como executar o mesmo procedimento com todos os vértices de destino, possa ser proporcionada a realização de uma única busca.

Foram realizadas buscas para cinco diferentes raios para cada cidade: 50, 100, 150, 200 e 250 metros; com dez mil pares de \gls{OD} totalizando 50.000 buscas para cada cidade, um total de 150.000 buscas por estratégia, somando 300.000 buscas. Essas buscas foram aglomeradas em 30 execuções de dez mil pares de \gls{OD} cada.

Além disso, os parâmetros citados acima buscam simular aspectos de cidades e pessoas reais, em que as buscas representam o movimento das pessoas dentro da cidade. Ademais, a variação dos valores de raio refere-se as pessoas que podem ter variações nas suas estruturas físicas. Por exemplo, idosos possuem um andar mais lentificado e com menor propensão para andar do que os adolescentes, estes, por sua vez, possuem uma maior disposição para realizar trajetos a longas distâncias, o que acarreta maiores possibilidades de variações de origens e destinos disponíveis nos itinerários. Ainda sobre esse assunto, ao afirmar sobre a quantidades de pares de \gls{OD}s, este se refere a representação das pessoas se locomovendo de todos os lugares das cidades.

Para executar os algoritmos de busca, foi utilizado o mesmo servidor para processar as duas estratégias para a mesma cidade, ou seja, a estratégia de colapsar executada na cidade de Fortaleza foi rodada na mesma máquina que o algoritmo de força bruta para todos os raios de busca. O código executado foi desenvolvido em C\# da plataforma \textit{dotnet core} versão 2.2 em ambiente Linux 18.04 LTS com as seguintes configurações de \textit{hardware} na Tabela \ref{tab:hardware-config}.

\begin{table}[h]
\centering
\caption{Configuração do servidor utilizado}
\label{tab:hardware-config}
\begin{tabular}{ l l l }
\hline
Dispositivo & \textit{Hardware}  & Descrição                                 \\ \hline
/0/0        & Memória     & 62GiB \textit{System memory}             \\ 
/0/1        & processador & Intel(R) Xeon(R) CPU E5-2630 v4 @ 2.20GHz \\ 
/0/2        & processador & Intel(R) Xeon(R) CPU E5-2630 v4 @ 2.20GHz \\ 
/0/3        & processador & Intel(R) Xeon(R) CPU E5-2630 v4 @ 2.20GHz \\ 
/0/4        & processador & Intel(R) Xeon(R) CPU E5-2630 v4 @ 2.20GHz \\ 
/0/5        & processador & Intel(R) Xeon(R) CPU E5-2630 v4 @ 2.20GHz \\ 
/0/6        & processador & Intel(R) Xeon(R) CPU E5-2630 v4 @ 2.20GHz \\ 
/0/7        & processador & Intel(R) Xeon(R) CPU E5-2630 v4 @ 2.20GHz \\ 
/0/8        & processador & Intel(R) Xeon(R) CPU E5-2630 v4 @ 2.20GHz \\ 
\hline
\end{tabular}
\end{table}

Dando continuidade às execuções, uma rotina, escrita em \textit{bash script} para Linux, permite que cada uma das 30 execuções sejam efetuadas automaticamente gerando dois arquivos de resultado para cada processamento. O primeiro arquivo de resultado diz respeito ao \textit{log} do programa com dados de duração, erros sistêmicos, estado de execução (Iniciado, em execução, erro, finalizado) e progresso dentre as 10.000 buscas. O segundo arquivo é o principal que contém as informações para cada \gls{OD} na ordem da lista ordenada a seguir:

\begin{enumerate}
  \item Origem, vértice de origem de onde foi traçado o raio para a região inicial
  \item Destino, vértice de destino de onde foi traçado o raio para a região final
  \item Status, se a busca encontrou o destino ou enfrentou algum problema
  \item Saltos, refere-se a quantos vértices formaram o caminho encontrado
  \item Expansões, o número de vértices que foram visitados até que o destino fosse encontrado
  \item Tempo, duração para que se realizasse a busca naquele par \gls{OD}
  \item Distância, o tamanho do caminho encontrado entre a origem e o destino
  \item Caminho, o caminho resultado, sendo o primeiro vértice o que foi selecionado na origem e o último vértice o que foi selecionado no destino
\end{enumerate}

Tendo em vista a configuração de \textit{hardware}, foram disponibilizados oito núcleos lógicos com aproximadamente 60 GB de memória \textit{RAM} para cada execução, sendo assim o algoritmo desenvolvido possui suporte a múltiplas \textit{threads} com uma estratégia de dividir o total de buscas para cada \textit{thread} com o resto, em caso de divisão não exata para a última \textit{thread}. Sendo assim, cada linha de execução realizou 1250 buscas e cada uma possuía sua própria instância do grafo da cidade que está em execução. Para o dado de tempo total, foi somado o tempo de cada busca para cada linha de execução podendo, por exemplo, realizar dois anos de busca em apenas três meses com a paralelização das buscas.

No contexto de validação do código, foram escritos 104 testes unitários e o versionamento foi garantido pela ferramenta \textit{Git}. O código do software desenvolvido está em \cite{danielaragao2019} e está disponível para download e futuras implementações.
	\chapter{Resultados}
\label{chap:resultados}
\begin{figure}[h]
    \centering
    \Caption{\label{fig:painel-tempo} Tempo de execução para cada cidade em função do raio.}	
    \UNIFORfig{}{
        \fbox{\includegraphics[width=16cm]{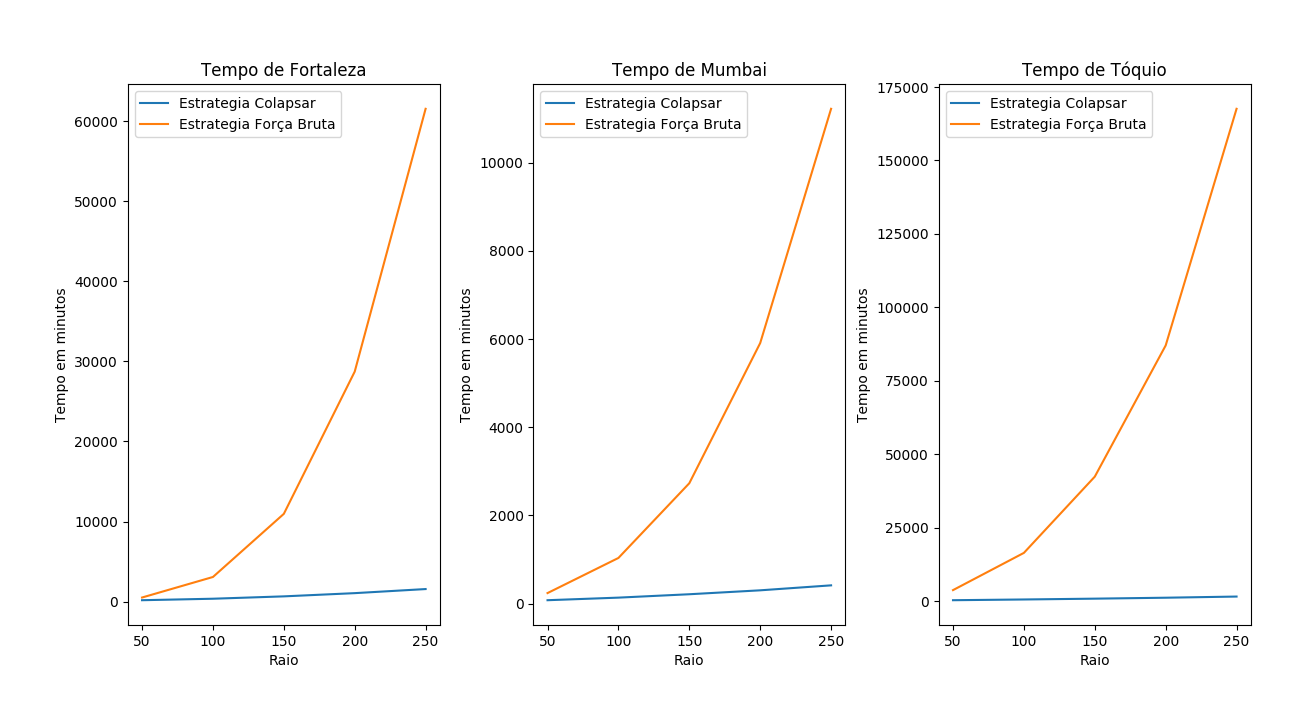}}
    }{
        \Fonte{Elaborado pelo autor}
    }
\end{figure}

\begin{figure}[h]
    \centering
    \Caption{\label{fig:painel-log-tempo} Logaritmo do tempo de execução para cada cidade em função do raio.}	
    \UNIFORfig{}{
        \fbox{\includegraphics[width=16cm]{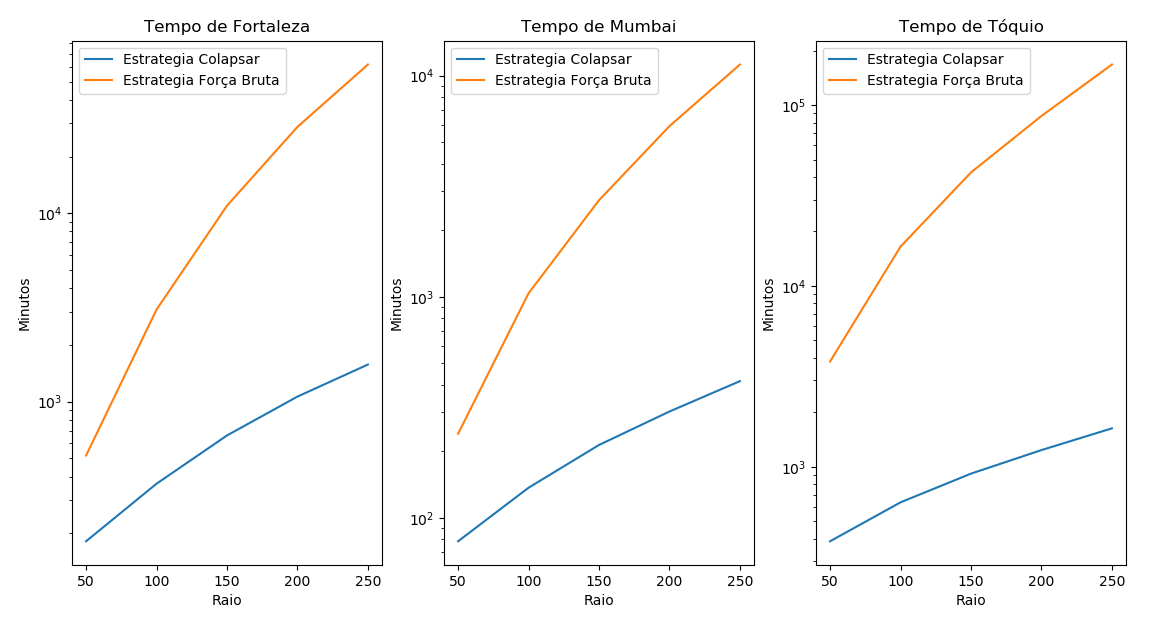}}
    }{
        \Fonte{Elaborado pelo autor}
    }
\end{figure}

Como resultado, foram geradas cinco métricas apresentadas em gráficos, para cada cidade. Para justificar o uso da estratégia de colapsar, como métrica principal foi escolhido o tempo de execução. Como dito no tópico anterior o registro do tempo foi emitido em cada caminho calculado, permitindo somar o tempo em um total para cada estratégia em cada cidade para cada raio. Pode-se observar os gráficos de cada cidade na Figura \ref{fig:painel-tempo} com o tempo de execução em minutos em função do raio e também na Figura \ref{fig:painel-log-tempo} com o logaritmo do tempo em função do raio.

Um comportamento padrão surgiu nas três cidades, podendo observar que para um raio não tão grande, como de 50 metros, não é possível ver muita diferença na utilização da estratégia. Porém ao iniciar o crescimento do raio nota-se que enquanto os dados de Colapsar se mantiveram quase constantes, os da Força Bruta cresceram de forma exponencial, onde para o maior raio, atingindo em Tóquio. A maior cidade em quantidade de vértices, onde o tempo teve uma duração de 121 dias, enquanto a estratégia de Colapsar precisou de somente 27 horas, como pode ser observado na Tabela \ref{tab:strategy-times}.

\begin{table}[h]
\centering
\caption{Tempo em horas e dias para execução de cada estratégia para cada cidade}
\label{tab:strategy-times}
\begin{tabular}{ c c c c }
\hline
Estratégia  & Fortaleza & Mumbai & Tóquio \\ 
\hline
Colapsar    & 26 h      & 7 h    & 27 h   \\
Força Bruta & 42 dias     & 8 dias  & 121 dias \\ 
\hline
\end{tabular}
\end{table}

\begin{figure}[h]
\centering
\Caption{\label{fig:painel-colapsados} Quantidade de nós colapsados por cidade.}	
\UNIFORfig{}{
 \fbox{\includegraphics[width=16cm]{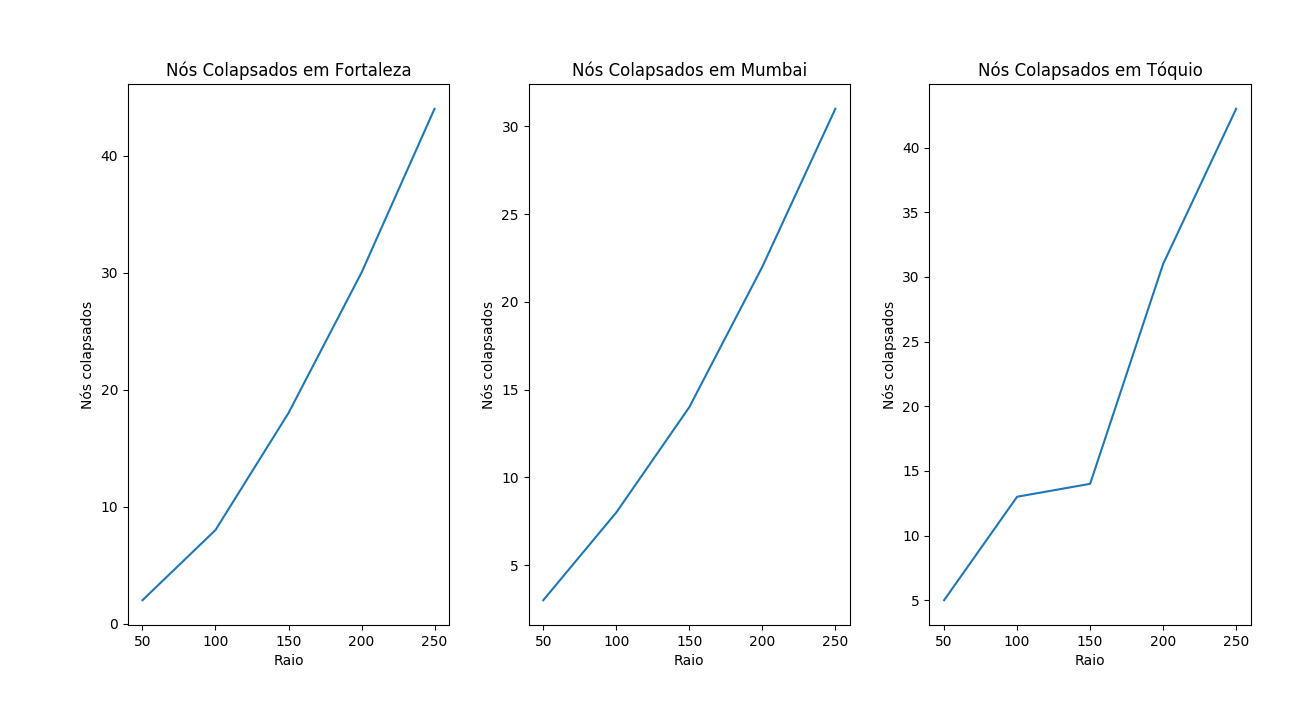}}
}{
 \Fonte{Elaborado pelo autor}
}
\end{figure}

\begin{figure}[h]
    \centering
    \Caption{\label{fig:painel-acuracia} Acurácia por cidade. Representando o percentual de caminhos ótimos formados}
    \UNIFORfig{}{
        \fbox{\includegraphics[width=16cm]{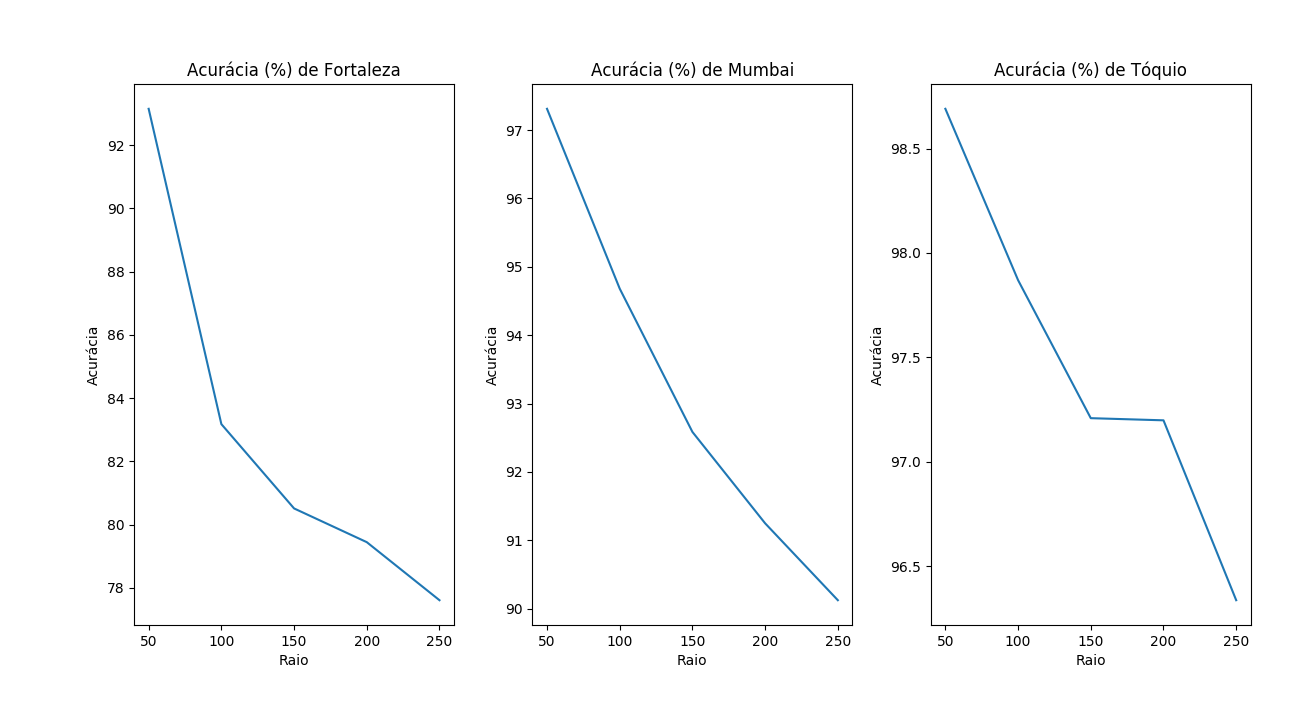}}
    }{
    \Fonte{Elaborado pelo autor}
    }
\end{figure}

Ainda a respeito do tempo, é notável que há uma relação direta entre a densidade do grafo e o aumento do tempo para a estratégia de Força Bruta, pois não só se aumenta o tamanho dos caminhos, mas a quantidade de nós colapsados para cada raio. Como se se pode observar na Figura \ref{fig:painel-colapsados}, houve uma semelhança entre Fortaleza e Tóquio no maior raio, porém mudou nos anteriores. Observando esse fator de quantidade de nós colapsados, tem-se que não há influência direta na acurácia, pois ao observar a mudança do raio 100 para 150 em que houve um baixo aumento de nós colapsados, não se pode verificar o mesmo comportamento de constância na acurácia apresentada na Figura \ref{fig:painel-acuracia}. Apesar disso foi verificada uma constância no gráfico em intervalos de raio diferentes.

Quanto a acurácia, nota-se que há uma grande queda para a cidade de Fortaleza, similar ao apresentado por Ponte \textit{et al} (2016), porém para as demais cidades não há uma queda tão grave, mostrando que o formato da rede influência diretamente nessa propriedade. Contudo o mesmo não pode ser dito para o tamanho, já que Mumbai, cidade menor, e Tóquio, cidade maior obtiveram resultados melhores.
 
\begin{figure}[h]
    \centering
    \Caption{\label{fig:painel-erro-maximo} Erro máximo por cidade.}	
    \UNIFORfig{}{
        \fbox{\includegraphics[width=16cm]{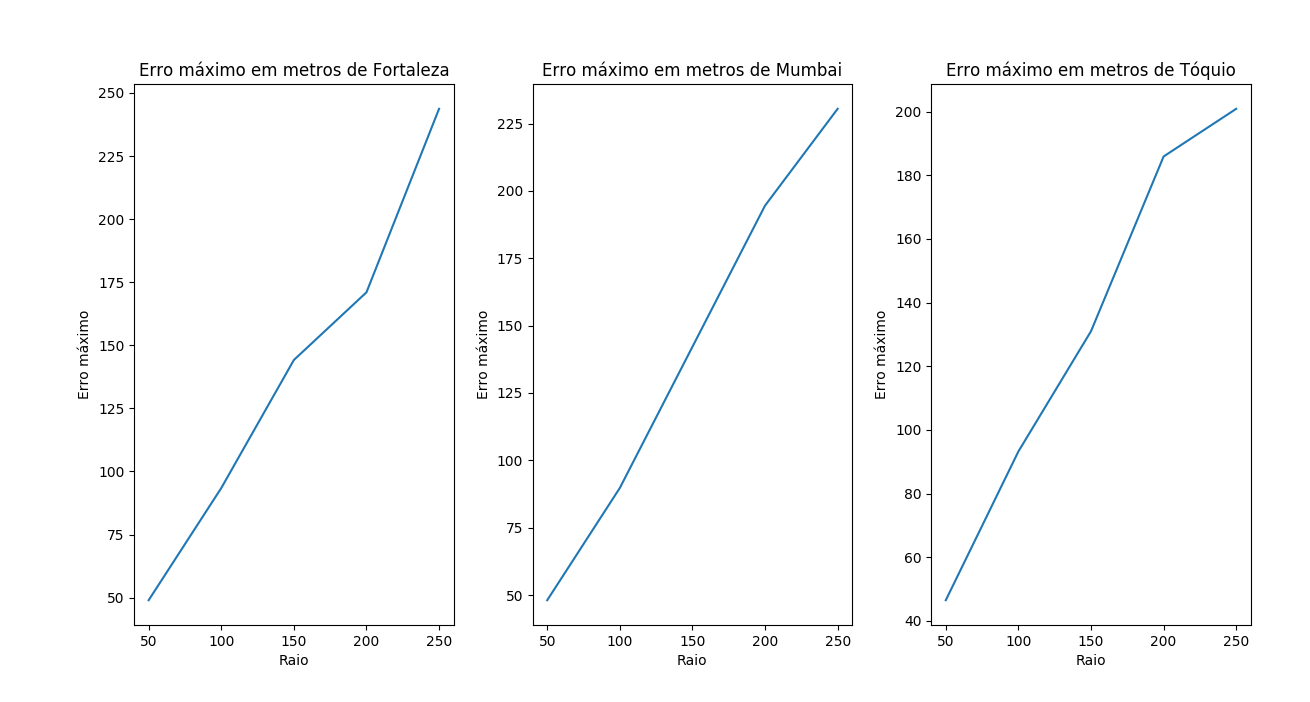}}
    }{
        \Fonte{Elaborado pelo autor}
    }	
\end{figure}

Falando de erro máximo, na Figura \ref{fig:painel-erro-maximo} nota-se que os caminhos encontrados tiveram como erro máximo um valor equivalente ao raio, ou seja, os erros vão ser proporcionais ao aumento da distância. Tornando o erro máximo previsível.
 
\begin{figure}[h]
    \centering
    \Caption{\label{fig:painel-erro-medio} Erro médio por cidade.}	
    \UNIFORfig{}{
        \fbox{\includegraphics[width=16cm]{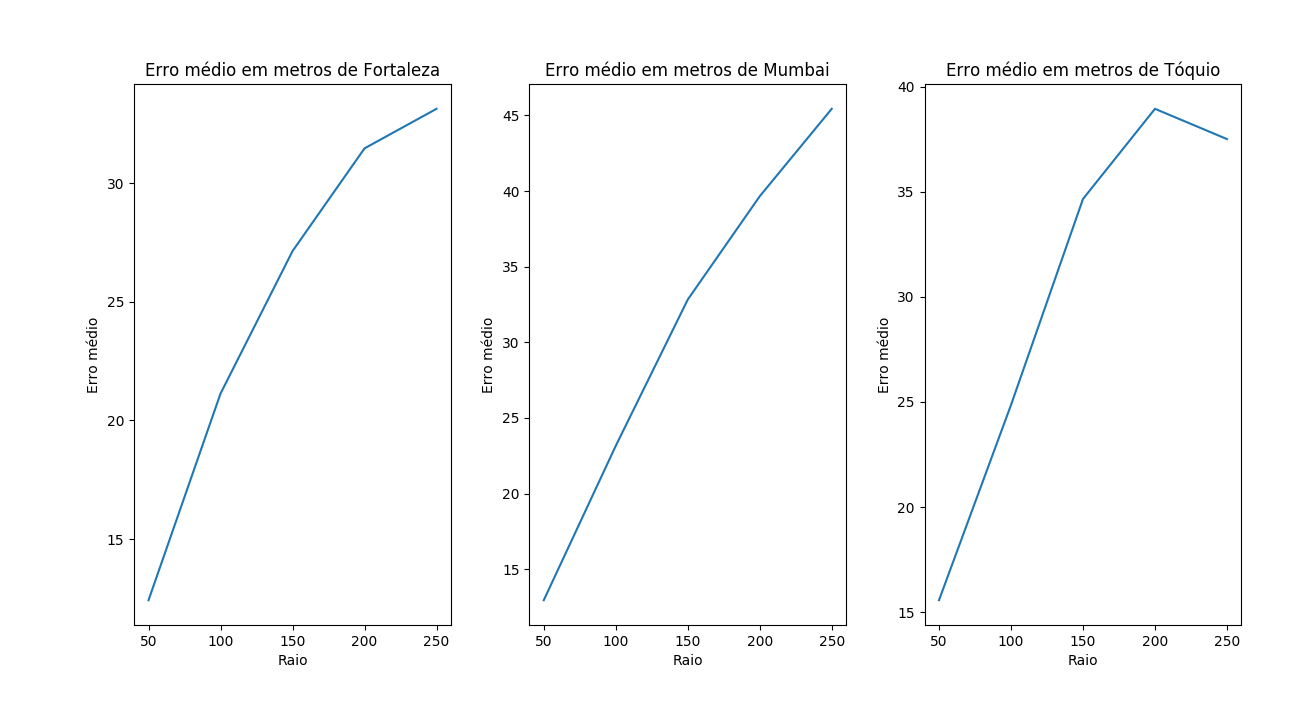}}
    }{
        \Fonte{Elaborado pelo autor}
    }	
\end{figure}

 Outro fator positivo para a generalização do modelo de Ponte \textit{et al} (2016), é que apesar de a estratégia não demonstrar resultados sempre ótimos, como é visto nos gráficos de acurácia, o erro médio da Figura \ref{fig:painel-erro-medio} não chega a ser maior que 50 metros, trazendo a situação em que os caminhos não ótimos encontrados possuem resultados quase ótimos. Já que para um contexto de cidades 50 metros não são valores significativos para quando se tem trajetos com quilômetros de distância para percorrer.

% \lipsum[2]

% \section{Resultados do Experimento A}
% \label{sec:resultados-do-experimento-a}

% \lipsum[3]

% \section{Resultados do Experimento B}
% \label{sec:resultados-do-experimento-b}

% \lipsum[4]
	\chapter{Conclusões e Trabalhos Futuros}
\label{chap:conclusoes-e-trabalhos-futuros}

%Para concluir, foi constatado a partir deste trabalho que não houve grande perda no uso da estratégia de colapsar em redes de grande porte como cidades reais e com topologia diferente da rede de ônibus de Fortaleza. Isso permite, então, adotar a estratégia de forma mais generalizada em grafos que talvez tenham grandes quantidades de vértices sem que se distancie muito da otimalidade.

A estratégia de colapsar vértices pra fazer só uma busca, mostrou que diminuiu consideravelmente o tempo de execução do algoritmo. Deste modo, constatou-se que os resultados obtidos por Ponte \textit{et al} (2016), no grafo da rede de ônibus de Fortaleza, foram estendidos nos resultados desse estudo. Além disso, observa-se que uma busca entre múltiplas origens e destinos são mais performáticas, tendo em vista um erro satisfatório que é menor que o raio de execução, o que permite inferir a existência de uma metodologia viável. Ademais, embora algumas buscas encontrarem algum erro, essas não passam de 90\% de acurácia para duas das cidades utilizadas, mesmo para Fortaleza que teve o menor resultado de acurácia, ainda encontra-se pequenos erros ao se analisar o contexto de cidades inteiras.

Além disso, deve ser salientado uma ressalva sobre a aplicação dessa estratégia, visto que o erro máximo permitido cresce em função ao raio. Já no cenário descrito nesse trabalho, é a distância que uma pessoa esta disposta a se locomover que interfere, tornando possível afirmar que, quando há crescimento do raio, há também aumento do erro. Por conta disso, é importante ter cautela e limitar o tamanho do raio. %Ademais, vale ressaltar que o erro médio mostra resulta em 40 metros aproximadamente em cada cidade.

%Uma das desvantagens observadas na Figura \ref{fig:painel-erro-maximo} foi quanto ao erro seguir de forma quase proporcional ao raio. Sendo assim a estratégia mostra-se viável somente até quando o erro aceitável puder atingir os mesmos valores do raio adotado para a busca. Para problemas que não exigem tanta precisão, também pode-se concluir que a média de erros é baixa, não chegando sequer a 50\% do raio adotado.

\section{Contribuições do Trabalho}
\label{sec:contribuicoes-do-trabalho}

%Dado o objetivo secundário, uma das contribuições do trabalho foi o entendimento sobre as propriedades de redes complexas e a qual topologia as cidades apresentadas mais se assemelham. Por consequência, também foi desenvolvido um programa capaz de gerar grafos que representassem as topologias citadas na pesquisa.

%Como contribuição também, foi possível descrever a própria generalização da estratégia de Colapsar, mesmo com a perda da otimalidade, não tendo um raio maior que o erro aceitável.

Os resultados deste estudo apontam que é possível estender a aplicação de Ponte \textit{et al} (2016) no grafo de ônibus de Fortaleza em outros cenários, sendo, então, admissível aplicá-la em cidades distintas com formatos e tamanhos diferentes que variam de 23 mil a 116 mil vértices.

\section{Limitações}
\label{sec:limitacoes}

%Quanto às limitações, para que pudesse haver um grupo controle, foi necessário executar a estratégia de Força Bruta em todas as cidades e para cada raio. Porém as grandes quantidades de vértices demonstraram-se muito custosas em relação ao tempo de execução para raios não tão grandes, aumentando cada vez mais e exigindo grande poder de processamento.

%Sendo assim, o principal fator limitante foi o processamento exigido para que fossem concluídas todas as buscas para a coleta de dados. Algumas execuções exigem meses de processamento mesmo com o processamento paralelizado em oito \textit{threads} de execução. Para contornar o problema cidades tiveram que ser retiradas do planejamento inicial. Para as cidades com muitos vértices se torna inviável a execução de buscas com o raio muito alto para o método de Força Bruta, apesar de ainda apresentar um tempo satisfatório para a estratégia de Colapsar.

%Procurando amenizar o problema do tempo tentou-se estimar as durações em cada cidade, porém com o crescimento exponencial não há muita precisão quanto a estimativa baseada em execuções passadas. No entanto uma aproximação se tornava mais real quando eram utilizados dados do próprio progresso da execução. Isso significa que somente tinha-se alguma previsão a respeito da configuração atual, com os raios e cidades selecionados e executados.

Em decorrência de limitações de tempo para a execução desse estudo, não foi possível avaliar raios maiores, como extensões de 500 ou 1000 metros, assim como outros que foram utilizados por Ponte \textit{et al} (2016). Visto que, as redes utilizadas possuem tamanhos maiores do que as redes do estudo citado acima, o que tornaria necessário um recurso computacional que pudesse abranger essa análise. 

\section{Trabalhos Futuros}
\label{sec:trabalhos-futuros}

Os estudos futuros poderão executar o mesmo procedimento para os grafos sintéticos utilizados. De forma que seja permitido comparar as propriedades das topologias, observando qual possui um melhor desempenho e qual topologia as cidades deveriam adotar para que possa executar melhor a estratégia de Colapsar. Ademais, poderão também contemplar a avaliação de maiores raios de execução, além de diferentes e maiores cidades das que foram utilizadas.

	%Elementos pós-textuais
	\bibliography{elementos-pos-textuais/referencias}

\providecommand{\abntreprintinfo}[1]{%
 \citeonline{#1}}
\setlength{\labelsep}{0pt}\begin{thebibliography}{}
\providecommand{\abntrefinfo}[3]{}
\providecommand{\abntbstabout}[1]{}
\abntbstabout{v-1.9.6 }

\bibitem[Albert e Barab{\'a}si 2002]{albert2002statistical}
\abntrefinfo{Albert e Barab{\'a}si}{ALBERT; BARAB{\'A}SI}{2002}
{ALBERT, R.; BARAB{\'A}SI, A.-L. Statistical mechanics of complex networks.
\textbf{Reviews of modern physics}, APS, v.~74, n.~1, p.~47, 2002.
Dispon{\'\i}vel em: \url{http://barabasi.com/f/103.pdf}.}

\bibitem[Antiqueira \textit{et al.} 2005]{antiqueira2005modelando}
\abntrefinfo{Antiqueira \textit{et al.}}{ANTIQUEIRA \textit{et al.}}{2005}
{ANTIQUEIRA, L.; NUNES, M. d. G.~V.; JR, O. O.; COSTA, L. d.~F. Modelando
  textos como redes complexas. In:  \textbf{Anais do III Workshop em Tecnologia
  da Informa{\c{c}}{\~a}o e da Linguagem Humana}. [s.n.], 2005. p. 22--26.
  Dispon{\'\i}vel em: \url{http://nilc.icmc.usp.br/til/til2005/arq0054.pdf}.}

\bibitem[Bart{\'a}k, Dovier e Zhou 2016]{bartak2016multiple}
\abntrefinfo{Bart{\'a}k, Dovier e Zhou}{BART{\'A}K; DOVIER; ZHOU}{2016}
{BART{\'A}K, R.; DOVIER, A.; ZHOU, N.-F. Multiple-origin-multiple-destination
  path finding with minimal arc usage: complexity and models. In:  IEEE.
  \textbf{2016 IEEE 28th International Conference on Tools with Artificial
  Intelligence (ICTAI)}. [S.l.], 2016. p. 91--97.}

\bibitem[Bast \textit{et al.} 2016]{bast2016route}
\abntrefinfo{Bast \textit{et al.}}{BAST \textit{et al.}}{2016}
{BAST, H.; DELLING, D.; GOLDBERG, A.; M{\"U}LLER-HANNEMANN, M.; PAJOR, T.;
  SANDERS, P.; WAGNER, D.; WERNECK, R.~F. Route planning in transportation
  networks. In:  \textbf{Algorithm engineering}. [S.l.]: Springer, 2016. p.
  19--80.}

\bibitem[Caminha e Furtado 2012]{caminha2012modeling}
\abntrefinfo{Caminha e Furtado}{CAMINHA; FURTADO}{2012}
{CAMINHA, C.; FURTADO, V. Modeling user reports in crowdmaps as a complex
  network. In:  \textbf{Proceedings of 21st International World Wide Web
  Conference. Citeseer}. [S.l.: s.n.], 2012.}

\bibitem[Caminha \textit{et al.} 2017]{caminha2017human}
\abntrefinfo{Caminha \textit{et al.}}{CAMINHA \textit{et al.}}{2017}
{CAMINHA, C.; FURTADO, V.; PEQUENO, T.~H.; PONTE, C.; MELO, H.~P.; OLIVEIRA,
  E.~A.; JR, J.~S. A. Human mobility in large cities as a proxy for crime.
\textbf{PloS one}, Public Library of Science, v.~12, n.~2, p. e0171609, 2017.}

\bibitem[Caminha \textit{et al.} 2016]{caminha2016micro}
\abntrefinfo{Caminha \textit{et al.}}{CAMINHA \textit{et al.}}{2016}
{CAMINHA, C.; FURTADO, V.; PINHEIRO, V.; SILVA, C. Micro-interventions in urban
  transportation from pattern discovery on the flow of passengers and on the
  bus network. In:  IEEE. \textbf{2016 IEEE International Smart Cities
  Conference (ISC2)}. [S.l.], 2016. p.~1--6.}

\bibitem[Chrispino \textit{et al.} 2013]{chrispino2013area}
\abntrefinfo{Chrispino \textit{et al.}}{CHRISPINO \textit{et al.}}{2013}
{CHRISPINO, A.; LIMA, L.~S. de; ALBUQUERQUE, M.~B. de; FREITAS, A. C.~C. de;
  SILVA, M. A. F.~B. da. A {\'a}rea cts no brasil vista como rede social: onde
  aprendemos?
\textbf{Ci{\^e}ncia \& Educa{\c{c}}{\~a}o}, Universidade Estadual Paulista,
  v.~19, n.~2, p. 455--479, 2013.
Dispon{\'\i}vel em:
  \url{https://dialnet.unirioja.es/servlet/articulo?codigo=5285680}.}

\bibitem[Feofiloff, Kohayakawa e Wakabayashi 2011]{feofiloff2011introduccao}
\abntrefinfo{Feofiloff, Kohayakawa e Wakabayashi}{FEOFILOFF; KOHAYAKAWA;
  WAKABAYASHI}{2011}
{FEOFILOFF, P.; KOHAYAKAWA, Y.; WAKABAYASHI, Y. Uma introdu{\c{c}}{\~a}o
  sucinta {\`a} teoria dos grafos.
2011.}

\bibitem[Filho 2019]{danielaragao2019}
\abntrefinfo{Filho}{FILHO}{2019}
{FILHO, D. A.~A. \textbf{Colapsar}. 2019.
Dispon{\'\i}vel em: \url{https://github.com/Daniel-Aragao/Colapsar\_cs}.}

\bibitem[GAIOSO, Paula \textit{et al.} 2013]{gaioso2013paralelizaccao}
\abntrefinfo{GAIOSO, Paula \textit{et al.}}{GAIOSO; PAULA \textit{et
  al.}}{2013}
{GAIOSO, R.; PAULA, L. \textit{et al.} Paraleliza{\c{c}}{\~a}o do algoritmo
  floyd-warshall usando gpu.
\textbf{SIMP{\'O}SIO EM SISTEMAS COMPUTACIONAIS. XIV}, 2013.}

\bibitem[Goldberg e Werneck 2005]{goldberg2005computing}
\abntrefinfo{Goldberg e Werneck}{GOLDBERG; WERNECK}{2005}
{GOLDBERG, A.~V.; WERNECK, R. F.~F. Computing point-to-point shortest paths
  from external memory. In:  \textbf{ALENEX/ANALCO}. [S.l.: s.n.], 2005. p.
  26--40.}

\bibitem[Hart, Nilsson e Raphael 1968]{hart1968formal}
\abntrefinfo{Hart, Nilsson e Raphael}{HART; NILSSON; RAPHAEL}{1968}
{HART, P.~E.; NILSSON, N.~J.; RAPHAEL, B. A formal basis for the heuristic
  determination of minimum cost paths.
\textbf{IEEE transactions on Systems Science and Cybernetics}, IEEE, v.~4,
  n.~2, p. 100--107, 1968.}

\bibitem[Henzinger, Krinninger e Nanongkai 2016]{henzinger2016deterministic}
\abntrefinfo{Henzinger, Krinninger e Nanongkai}{HENZINGER; KRINNINGER;
  NANONGKAI}{2016}
{HENZINGER, M.; KRINNINGER, S.; NANONGKAI, D. A deterministic almost-tight
  distributed algorithm for approximating single-source shortest paths. In:
  ACM. \textbf{Proceedings of the forty-eighth annual ACM symposium on Theory
  of Computing}. [S.l.], 2016. p. 489--498.}

\bibitem[Karypis e Kumar 1995]{karypis1995metis}
\abntrefinfo{Karypis e Kumar}{KARYPIS; KUMAR}{1995}
{KARYPIS, G.; KUMAR, V. Metis-unstructured graph partitioning and sparse matrix
  ordering system.(1995).
\textbf{University of Minnesota, Department of Computer Science and
  Engineering}, 1995.
Dispon{\'\i}vel em: \url{https://dm.kaist.ac.kr/kse625/resources/metis.pdf}.}

\bibitem[Lewis 2009]{lewis.c2.a2009}
\abntrefinfo{Lewis}{LEWIS}{2009a}
{LEWIS, T.~G. Graphs.
In:  \underline{\ \ \ \ \ \ \ \ }. \textbf{Network science: theory and
  practice}. Haboken, New Jersey: John Wiley \& Sons, Inc., 2009. cap.~2.}

\bibitem[Lewis 2009]{lewis.c6.a2009}
\abntrefinfo{Lewis}{LEWIS}{2009b}
{LEWIS, T.~G. Scale-free network.
In:  \underline{\ \ \ \ \ \ \ \ }. \textbf{Network science: theory and
  practice}. Haboken, New Jersey: John Wiley \& Sons, Inc., 2009. cap.~6.}

\bibitem[Newman 2010]{newman2010}
\abntrefinfo{Newman}{NEWMAN}{2010}
{NEWMAN, M. E.~J. \textbf{Networks: An Introduction}. New York: Oxford
  University Press, 2010.}

\bibitem[Oliveira \textit{et al.} 2012]{oliveira2012monitoramento}
\abntrefinfo{Oliveira \textit{et al.}}{OLIVEIRA \textit{et al.}}{2012}
{OLIVEIRA, R.; ARA{\'U}JO, J.; MEDEIROS, F.; BRITO, A. Monitoramento das
  intera{\c{c}}{\~o}es dos aprendizes na rede social twitter como apoio ao
  processo de media{\c{c}}{\~a}o docente. In:  \textbf{Brazilian Workshop on
  Social Network Analysis and Mining, BrasNAM}. [S.l.: s.n.], 2012. p.
  130--148.}

\bibitem[Ponte, Caminha e Furtado 2016]{ponte2016busca}
\abntrefinfo{Ponte, Caminha e Furtado}{PONTE; CAMINHA; FURTADO}{2016}
{PONTE, C.; CAMINHA, C.; FURTADO, V. Busca de melhor caminho entre dois pontos
  quando m{\'u}ltiplas origens e m{\'u}ltiplos destinos s{\~a}o poss{\'\i}veis.
\textbf{Recife: ENIAC}, 2016.}

\bibitem[Ravasz e Barab{\'a}si 2003]{ravasz2003hierarchical}
\abntrefinfo{Ravasz e Barab{\'a}si}{RAVASZ; BARAB{\'A}SI}{2003}
{RAVASZ, E.; BARAB{\'A}SI, A.-L. Hierarchical organization in complex networks.
\textbf{Physical review E}, APS, v.~67, n.~2, p. 026112, 2003.}

\bibitem[Russel 2004]{norvig2004}
\abntrefinfo{Russel}{RUSSEL}{2004}
{RUSSEL, P.~N. S. Além da busca clássica.
In:  \underline{\ \ \ \ \ \ \ \ }. \textbf{Inteligência Artificial}. Rio de
  Janeiro: Elsevier Editora Ltda, 2004.  (3), cap.~4.}

\bibitem[Sayama 2015]{sayama2015}
\abntrefinfo{Sayama}{SAYAMA}{2015}
{SAYAMA, H. Introduction.
In:  \underline{\ \ \ \ \ \ \ \ }. \textbf{Introduction to the Modeling and
  Analysis of Complex Systems}. Open SUNY Textbooks, 2015. cap.~1.
Dispon{\'\i}vel em:
  \url{https://textbooks.opensuny.org/introduction-to-the-modeling-and-analysis-of-complex-systems/}.}

\bibitem[Sullivan \textit{et al.} 2017]{sullivan2017towards}
\abntrefinfo{Sullivan \textit{et al.}}{SULLIVAN \textit{et al.}}{2017}
{SULLIVAN, D.; CAMINHA, C.; MELO, H.~P.; FURTADO, V. Towards understanding the
  impact of crime on the choice of route by a bus passenger. In:  SPRINGER.
  \textbf{EPIA Conference on Artificial Intelligence}. [S.l.], 2017. p.
  41--50.}

\bibitem[Vajapeyam 2014]{vajapeyam2014}
\abntrefinfo{Vajapeyam}{VAJAPEYAM}{2014}
{VAJAPEYAM, S. Understanding shannon's entropy metric for information.
\textbf{arXiv preprint arXiv:1405.2061}, 2014.
Dispon{\'\i}vel em: \url{https://arxiv.org/abs/1405.2061 25/04/2019}.}

\bibitem[Wang e Chen 2003]{wang2003complex}
\abntrefinfo{Wang e Chen}{WANG; CHEN}{2003}
{WANG, X.~F.; CHEN, G. Complex networks: small-world, scale-free and beyond.
\textbf{IEEE circuits and systems magazine}, IEEE, v.~3, n.~1, p. 6--20, 2003.
Dispon{\'\i}vel em:
  \url{http://cs.engr.uky.edu/~silvestri/teaching/complexNetworks/papers/Wang-and-Chen-Small-world-and-Scale-Free.pdf}.}

\end{thebibliography}
	%\imprimirglossario
	%\imprimirapendices
		% Adicione aqui os apendices do seu trabalho
		%\input{elementos-pos-textuais/apendices/historico-de-mudancas}
		%\input{elementos-pos-textuais/apendices/lorem-ipsum}
		%\input{elementos-pos-textuais/apendices/termo-de-fiel-depositario}
	%\imprimiranexos
		% Adicione aqui os anexos do seu trabalho
		%\input{elementos-pos-textuais/anexos/exemplo-de-anexo}
		%\input{elementos-pos-textuais/anexos/dinamica-das-classes-sociais}
	\imprimirindice

\end{document}